\documentclass[runningheads]{llncs}

\usepackage{eccv}

\usepackage{eccvabbrv}

\usepackage{graphicx}
\usepackage{booktabs}

\usepackage[accsupp]{axessibility}  

\usepackage{hyperref}

\usepackage{orcidlink}
\usepackage{makecell}
\newcommand{\repeatthanks}{\textsuperscript{\thefootnote}}

\begin{document}

\title{Adapt2Reward: Adapting Video-Language Models to Generalizable Robotic Rewards via Failure Prompts} 

\titlerunning{Adapt2Reward}

\author{Yanting Yang \inst{1}\thanks{Equal contribution;$^{\dagger}$ Corresponding author}  \and 
Minghao Chen\inst{2}\repeatthanks$^{\dagger}$ \and
Qibo Qiu\inst{3,7}\and
Jiahao Wu\inst{4} \and
Wenxiao Wang\inst{1} \and
Binbin Lin\inst{1,5} \and
Ziyu Guan\inst{6} \and
Xiaofei He\inst{7}
}

\authorrunning{Y.~Yang et al.}

\institute{$^{1}$School of Software Technology, Zhejiang University, \\
$^{2}$School of Computer Sciene and Technology, 
 Hangzhou Dianzi University,\\$^{3}$
China Mobile (Zhejiang) Research \& Innovation Institute, \\$^{4}$
 The Hong Kong Polytechnic University, $^{5}$ Zhiyuan Research Institute, \\$^{6}$
 School of Computer Sciene and Technology, Xidian University,\\$^{7}$
 State Key Lab of CAD\&CG, Zhejiang University
 \\ {\tt\small \{yantingyang,qiuqibo\_zju,wenxiaowang,binbinlin\}@zju.edu.cn,}
 \\ {\tt\small minghaochen01@gmail.com, jiahao.wu@connect.polyu.hk,}
  \\ {\tt\small zyguan@xidian.edu.cn, xiaofeihe@cad.zju.edu.cn} 
}

\maketitle

\begin{abstract}
For a general-purpose robot to operate in reality, executing a broad range of instructions across various environments is imperative. Central to the reinforcement learning and planning for such robotic agents is a generalizable reward function. Recent advances in vision-language models, such as CLIP, have shown remarkable performance in the domain of deep learning, paving the way for open-domain visual recognition. However, collecting data on robots executing various language instructions across multiple environments remains a challenge. This paper aims to transfer video-language models with robust generalization into a generalizable language-conditioned reward function, only utilizing robot video data from a minimal amount of tasks in a singular environment. Unlike common robotic datasets used for training reward functions, human video-language datasets rarely contain trivial failure videos. To enhance the model's ability to distinguish between successful and failed robot executions, we cluster failure video features to enable the model to identify patterns within. For each cluster, we integrate a newly trained failure prompt into the text encoder to represent the corresponding failure mode. Our language-conditioned reward function shows outstanding generalization to new environments and new instructions for robot planning and reinforcement learning. 
\end{abstract}
\section{Introduction}
\label{sec:intro}

Recent research in the field of ``generalist robots''~\cite{CLIPort,BC-Z,RT-1,Stone2023OpenWorldOM} aims to equip robots with the ability to perform a wide range of tasks conditioned on natural language instructions. This emerging field is propelled by breakthroughs in multi-modal models, such as CLIP~\cite{CLIP}, BLIP~\cite{BLIP}, and DALLE-2~\cite{dall-e2}, known for their exceptional proficiency in vision-language tasks. Their success stems from training on comprehensive datasets encompassing extensive human data across varied environments. Inspired by these achievements, some notable works in robotics~\cite{BC-Z,RT-1,Stone2023OpenWorldOM,RT-2} have employed rich and diverse robotic interaction datasets. However, the collection of such comprehensive, high-quality robotic data is considerably more challenging and labor-intensive than collecting datasets for computer vision and natural language processing~\cite{CLIP,LAION-400M,LAION-5B,Xu2024HeterogeneousDF,Wu2023LeveragingLL}.

\begin{figure}[t]
\centering
\includegraphics[width=0.86\textwidth]{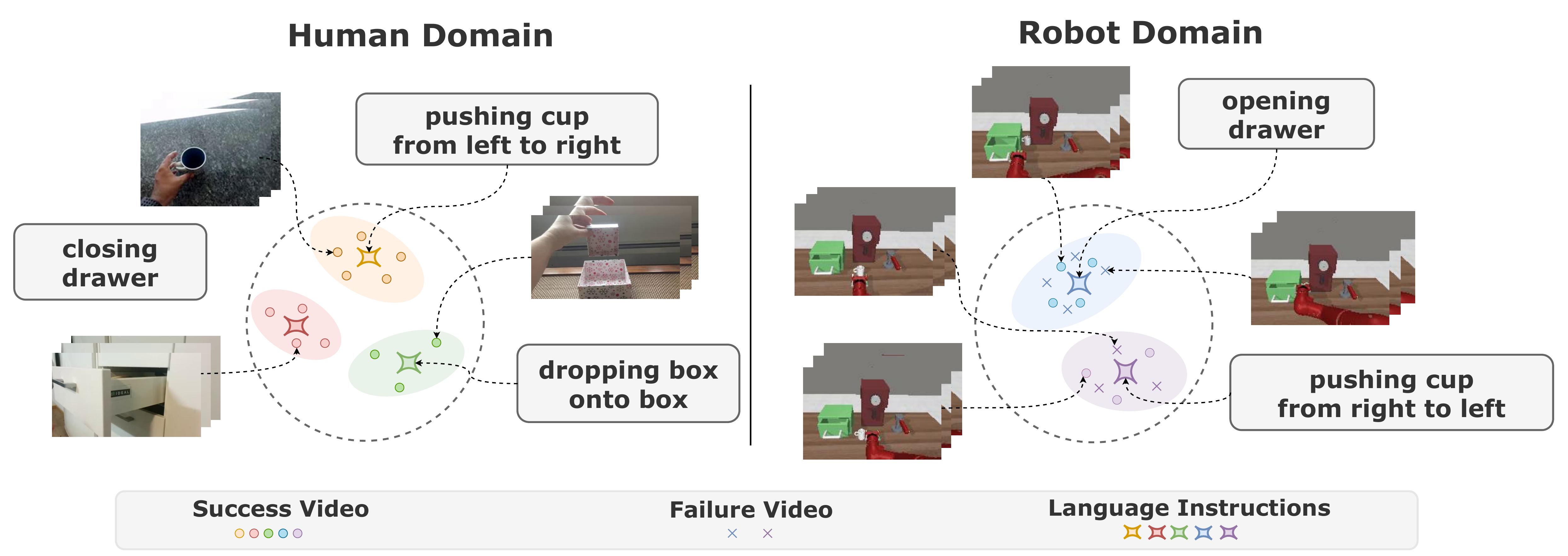}
\caption{
Human video-language datasets typically lack failure videos. This limitation will result in models that are effective at categorizing tasks but exhibit diminished efficiency in distinguishing between successful and unsuccessful task executions.}
\label{fig1}
\end{figure}

One promising approach to tackling these challenges is developing a generalized reward function, which can assess the success of robotic behaviors. Traditional methods, e.g., inverse reinforcement learning~\cite{irl2004}, often result in specialized reward functions limited to a narrow range of tasks. In this paper, we attempt to adapt pre-trained vision-language models as reward functions. Such models potentially offer broad generalization~\cite{R3M,Radosavovic2022RealWorldRL}. However, applying these models to robotic reward learning is not straightforward. Firstly, due to the scarcity of robotic interaction data, fine-tuning these models on limited robotic datasets risks model overfitting and catastrophic forgetting, undermining their generalization ability. Secondly, there's a significant domain shift and embodiment gap when transitioning from human-centric data to robotic applications. Unlike human videos, which feature diverse environments and perspectives, robot videos typically originate from more controlled, static settings, and the mechanical nature of robotic movements differs markedly from the fluidity of human actions. Previous work, i.e., DVD~\cite{DVD}, has attempted to address these challenges by training a discriminator to classify if two videos perform the same task. However, this approach overlooks the need for a binary classification of success and failure in reward functions, underscoring the necessity for data on failed robotic tasks.

In contrast to typical robotic datasets used for training reward functions~\cite{LOReL}, human video-language datasets rarely include inconsequential or minor failure videos. This discrepancy often leads to models proficient in categorizing tasks but less capable of distinguishing successful from unsuccessful executions. As Fig.~\ref{fig1} shows, while a video-language model can effectively distinguish the ``closing a drawer'' task from the ``opening a drawer'' task, it may not identify failure videos in the task, such as incomplete closure or accidental reopening. Consequently, training solely with successful videos is insufficient for the model to recognize and classify diverse failure scenarios accurately. A straightforward solution is to employ Binary Cross-Entropy (BCE) loss in training to enhance the distinction between success and failure samples. However, this approach can be problematic. The model can assign high scores to success samples in the training set and underscore others, leading to poor performance in novel situations and tasks.

To address these challenges, we introduce a novel method integrating learnable failure prompts within the model architecture. This approach is based on the hypothesis that task failures can be grouped into several modes and potentially transferable across similar tasks. We begin by clustering failure videos to identify distinct failure patterns, assigning each cluster a unique identifier for a corresponding failure prompt. This strategy enables the model to develop a nuanced understanding of different failure modes, facilitating knowledge transfer through learnable failure prompts and improving initial performance on unseen similar tasks. Furthermore, we incorporate cross-domain contrastive learning and domain-specific prompt learning to align text and video representations across human and robot domains. 

Our contributions can be summarized as follows:
\begin{itemize}
    \item We highlight the significance of including failed robotic videos in reward learning with human videos, and we identify the gap between human video-language datasets and robotic datasets in reward learning.
    \item We propose learnable failure prompts to model patterns of robotic failures effectively. The introduction of these prompts significantly enhances the model's adaptability and applicability.
    \item When combined with Visual Model Predictive Control (VMPC), our approach demonstrates superior generalization in the MetaWorld environment, outperforming previous methods. The effectiveness of our reward model for reinforcement learning is also illustrated in the Concept2Robot environment, where it generalizes to unseen tasks with varied viewpoints and objects.
\end{itemize}

\section{Related Works}
\label{sec:related}

\subsection{Reward Learning}
Reinforcement learning (RL) presents a dynamic framework for automating decision-making and control, though it often entails significant engineering of features and rewards for practical use. To address these challenges, inverse reinforcement learning (IRL)~\cite{irl2004} aims to infer experts' reward functions from observed behaviors, as extensively discussed in the literature~\cite{Jain2006MaximumMP,Ziebart2008MaximumEI,Wulfmeier2015MaximumED,Finn2016GuidedCL,Fu2018LearningRR,Xu2023PersonalizedRR}. Despite expansions to scenarios involving human-provided outcomes or demonstrations~\cite{Fu2018VariationalIC,Singh2019EndtoEndRR,Zakka2021XIRLCI,das2021model,Li2024ScenarioDrivenCS,Wu2022DisentangledCL}, existing research tends to focus on single tasks within confined environments, limiting wider application. Modern research trends are moving towards developing multi-task reward functions using visual inputs~\cite{DVD,Concept2Robot,LOReL,R3M}. An exemplar is LORel~\cite{LOReL}, which learns language-driven skills from sub-optimal offline data and crowd-sourced annotations. However, obtaining high-quality robot demonstrations remains resource-intensive. In contrast, our approach leverages extensive human-centric datasets, circumventing the need for robot hardware and capitalizing on the availability of online resources.

\subsection{Robotic Learning from Human Video}
Extensive research has been devoted to learning robotic behaviors from human videos. A prevalent method involves transforming human trajectories into robotic motions~\cite{Lee2013ASA,Yang2015RobotLM,Nguyen2017TranslatingVT,Rothfuss2018DeepEM,Lee2017LearningRA,Shaw2022VideoDexLD,Wang2023MimicPlayLI}, achieved by identifying and tracking hand positions in human videos and subsequently aligning them with corresponding robotic actions or primitives for task execution. Another tactic is pixel-level transfer, where human demonstrations or goals are directly transposed into a robotic framework~\cite{Liu2017ImitationFO,Smith2019AVIDLM,Sharma2019ThirdPersonVI}, employing both paired and unpaired datasets. 
Recent research has refined self-supervised algorithms to produce embeddings that are sensitive to object interactions and postures, while minimizing the impact of non-essential factors such as viewpoint and embodiment~\cite{R3M,Zakka2021XIRLCI,Xiao2022MaskedVP,Parisi2022TheUE,Sermanet2017TimeContrastiveNS}. 
Furthermore, recent advancements have recognized platforms like YouTube as substantial sources of ``in-the-wild'' visual data, featuring diverse human interactions, such as Something-Something-v2~\cite{something-something-v2}. Robots that can assimilate and learn reward functions from this extensive array of data have the potential for broad generalization. For instance, Concept2Robot~\cite{Concept2Robot} utilizes a pre-trained video classifier to infer robot reward functions. Similarly, DVD~\cite{DVD} proposes a domain-agnostic video discriminator, aimed at facilitating generalizable reward learning. LIV~\cite{liv} inducing a cross-modal embedding with temporal coherence and semantic alignment for language-image reward. 



\subsection{Language-Conditioned Robotic Learning}
Recently, CLIP~\cite{CLIP} aligns vision and language features from millions of image-caption pairs sourced from the internet, which is a robust foundation established for grounding semantic concepts prevalent across tasks. Therefore, prior research has predominantly concentrated on end-to-end learning of intricate robot manipulation, leveraging the multi-environment and multi-task robotic datasets, such as BC-Z~\cite{BC-Z}, CLIPort~\cite{CLIPort}, and RT-1~\cite{RT-1}. 
Recent studies~\cite{vip,MineDojoBO,VLMasSuccess,Kwon2023RewardDW} have also advocated for leveraging pretrained foundation models to generate reward.
Diverging from these approaches, most recent studies have utilized Large Language Models (LLMs) for automated generation of dense~\cite{Text2Reward, Eureka} or sparse~\cite{Yu2023LanguageTR} reward coding for policy learning or action synthesis in robotics. These methods, however, require extremely Large Language Models, i.e., GPT-4~\cite{OpenAI2023GPT4TR}, to generate accurate code for each task and are more suited for complex tasks requiring deep semantic understanding.
Our work deviates from previous efforts by circumventing the challenges of collecting diverse robotic data and the requirements of super-large models.



\section{Methods}


\label{sec:methods}

\begin{figure*}[t]
\centering
\includegraphics[scale=0.45]{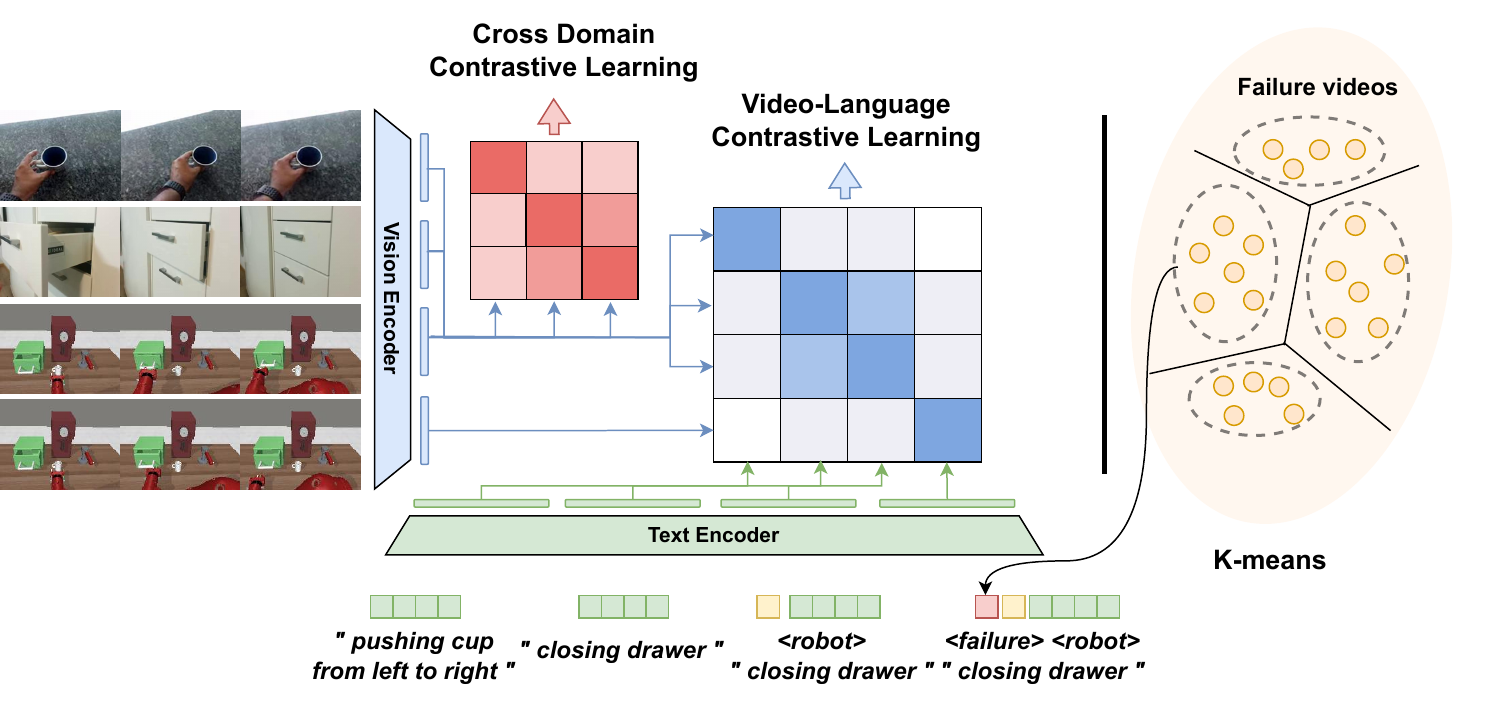}
\caption{\textbf{Adapt2Reward Architecture.} 
We propose Adapt2Reward which incorporates learnable failure prompts into the model's architecture.  Our approach starts with clustering failure videos to discern specific patterns. Each identified cluster is then associated with a unique failure prompt.  Additionally, we employ cross-domain contrastive learning and domain-specific prompt learning to align text and video representations between human and robot domains.}
\label{fig2}
\end{figure*}

\subsection{Preliminaries}
\textbf{Problem Statement.} In the setting of our problem, we consider a robotic agent to accomplish tasks from a task distribution, $\mathcal{T}\sim\mathcal{D}_{\mathcal{T}}$. Each task has some underlying reward function $R$ and can be expressed in natural language $l$. Consequently, for a specified task $\mathcal{T}$, our robotic agent operates in a fixed horizon Markov decision process: $(\mathcal{S}, \mathcal{A}, p, \mathcal{R}, T)$, where $\mathcal{S}$ denoted as the state space (image or short video clip in our case), $\mathcal{A}$ is the action space of the robot, $p(s_{t+1}|s_{t}, a_{t})$ represents the stochastic dynamics of robotic environments, $\mathcal{R}$ displays the reward for task $\mathcal{T}$, and $T$ is the episode horizon.
Our target is to learn a parametric model $\mathcal{R}_{\theta}$ that estimates the underlying reward function $\mathcal{R}$ for each task $\mathcal{T}$, conditioned on the natural language expression $l$. 
With such a reward function, we can utilize open-loop planners or reinforcement learning to optimize the reward for each task.

In order to learn a broadly generalized reward function $\mathcal{R}_{\theta}$, we have access to human dataset $\{(x^h_i, l^h_i)\}_{i=1}^N$, consisting of $N$ videos of human demonstration for various tasks, where $l^h_i$ is a sentence expressing a task $\mathcal{T}$ sampled from the human task distribution $\mathcal{T}\sim\mathcal{D}^{h}_{\mathcal{T}}$.
We are also given a finite robot dataset $\{(x^r_j,l^r_j,r_j)\}^{M}_{j=1}$ of videos of the robot doing tasks $\mathcal{T}\sim\mathcal{D}^{r}_{\mathcal{T}}$, in which the robot can complete the task, $r_j=1$, or the robot failed to finish the task, $r_j=0$. Although human data is widely available, robot data consists of only a limited number of tasks in a handful of environments, $\mathcal{D}^{r}_{\mathcal{T}}\subset\mathcal{D}_{\mathcal{T}}$. Therefore, we have many more diverse human video demonstrations than robot video demonstrations per task and many more tasks that have human videos but not robot videos, in which case $\mathcal{D}^{r}_{\mathcal{T}}\subset\mathcal{D}_{\mathcal{T}}\subset\mathcal{D}^{h}_{\mathcal{T}}$. We only use visual observations to determine whether a task is completed or not, and do not use low-dimensional states or actions. What is clear is that we will also face a large domain shift in the human and robot domains. By learning such a generalizable and robust reward model, we expect it to be able to complete these tasks in new environments and to generalize to unseen tasks.

\textbf{VLM as reward model.} Learning multi-modal representations using large-scale video-text pretraining has proved to be effective for a wide range of uni-modal and multi-modal applications that allows us to train multi-task success detectors by directly leveraging powerful pretrained Vision Language Model (VLM), such as CLIP4Clip~\cite{CLIP4Clip} and Singularity~\cite{singularity}.
In our work, the VLM is given a visual input representing the state of the world (a short video clip $x_{i}\in\mathbb{R}^{\mathit{L}\times\mathit{H}\times\mathit{W}\times\mathit{C}}$ with $\mathit{L}$ frames), and a corresponding text description $l_{i}$. The vision encoder encodes the $\mathit{T}$ frames as a batch of images independently and feeds these frame-level features to a temporal vision encoder to obtain fine-grained temporal video representation $\mathit{v}_{i}\in \mathbb{R}^{\mathit{D}}$. For the text describing the desired behavior or task, we denote the representation as  $\mathit{t}_{i}\in \mathbb{R}^{\mathit{D}}$. The video and text features are input to the multimodal encoder to get a video-level prediction score, which can serve as a reward.

\subsection{Human-Robot Contrastive learning}
\label{sec:3.2}

\textbf{Cross Domain Contrastive Learning.}
We present our methodology for leveraging contrastive learning to learn domain-invariant features by forming pairs across domains. Specifically, we hope that samples within the same category, irrespective of their domain origin, are positioned closely in feature space, while those from distinct classes are separated regardless of domain. More formally, we consider $\mathit{l}_2$-normalized features $v_{i}^{r}$ from the i-th sample $(x^r_i, l^r_i)$ in the robot domain as an anchor, and it forms a positive pair with the sample having the same expression from the human domain and robot domain, whose features are denoted as $v_{p}$. We formulate the cross-domain contrastive loss (CDC) as:
\begin{small} 
\begin{equation}
\mathcal{L}_{C D C}^{r, i}=-\frac{1}{|P(l^{r}_{i})|} \sum_{p \in P(l^{r}_{i})} \log \frac{\exp (\boldsymbol{v}_i^{r^{\top}} \boldsymbol{v}_p / \tau)}{\sum_{j=1}^B \exp (\boldsymbol{v}_i^{r^{\top}} \boldsymbol{v}_j / \tau)}
\label{H}
\end{equation}
\end{small}
where $\mathit{P}(l^{r}_{i}) = \{k | l^{h}_{k}= l^{r}_{i}, k\in\{1,..,B_h\}\}\cup\{k | l^{r}_{k} = l^{r}_{i}, k\in\{1,..,B_r\}\}$ indicates the set of positive samples from the cross-domain that share the same label with the anchor $x_{i}^{r}$ and $B=B_h+B_r$ denotes the batch size. In each batch, we sample $B_{h}$ human samples from the human dataset, and $B_{r}$ successful robot samples ($r_j=1$) from the robot dataset.
In Eqn. \ref{H}, we consider samples from the robot domain as anchors. Alternatively, we can use human samples as anchors and compute $\mathcal{L}_{C D C}^{h, i}$
Then, we combine $\mathcal{L}_{C D C}^{r, i}$ with $\mathcal{L}_{C D C}^{h, i}$ to derive the cross-domain contrastive loss as follows:
\begin{equation}
\mathcal{L}_{C D C}=\sum_{i=1}^{B_h} \mathcal{L}_{C D C}^{h, i}+\sum_{i=1}^{B_r} \mathcal{L}_{C D C}^{r, i} 
\end{equation}

\textbf{Video-Language Contrastive Learning.} 
To promote the model's ability to capture semantically pertinent features across human and robot domains, we employ a video-language contrastive loss. This approach, distinct from conventional video-language alignment, aims to minimize discrepancies in both domains. It not only aligns temporal dynamics with task semantics but also enhances the adaptation of video features by leveraging the shared action semantics across both domains. Formally, we denote the video-text paired features $(v_i, t_i)$ from the human or robot domain, where $v_i \in \left\{v^{r}_i, v^{h}_i\right\}$ and $t_i \in \left\{t^{r}_i, t^{h}_i\right\}$. The modified video-language contrastive loss (VLC) is defined as:
\begin{equation}
\mathcal{L}_{VLC}^{v_i, t_i}=\frac{\exp \left(v^{\top}_i t_i / \tau\right)}{\sum^B_{j=1} \exp \left(v^{\top}_i t_j / \tau\right)}
\label{HH}
\end{equation}
where $\tau$ is the temperature. In particular, we minimize the sum of two multimodal contrastive losses:
\begin{equation}
\mathcal{L}_{V L C}=\sum_{i=1}^{B} \mathcal{L}_{V L C}^{v_i, t_i}+\sum_{i=1}^{B} \mathcal{L}_{V L C}^{t_i, v_i} 
\end{equation}

\subsection{Learning from failure}

\label{sec:3.3}


In human cognition, task acquisition often hinges on introspective analysis of failures, leading to insights into causal missteps. Consequently, we posit that the integration of robot failure data into reward learning can enrich the model's capability to distinguish between efficacious and errant actions. 

\textbf{Binary Cross-Entropy Loss.} To distinguish between successful and failed videos, we adopt Binary Cross-Entropy (BCE) loss, a methodologically straightforward yet effective approach. Concretely, for each training batch, apart from $B_h$ human samples and $B_r$ successful robot samples, we sample $B_f$ failed robot samples ($r_i=0$) from the robot dataset. We denote $\mathit{l}_2$-normalized features $(v_{i}^{r}, t_{i}^{r})$ from a sample $(x_{i}^{r}, l_{i}^{r}, r_i)$ in successful or failed robot samples. We then minimize the binary cross-entropy loss:
\begin{equation}
\mathcal{L}_{B C E}=-\sum_{i=1}^{B_r+B_f} [r_i\log(p_i) +(1-r_i) (1-\log(p_i)))]
\end{equation}
where $p_i = \sigma({\boldsymbol{v}_i^r}^{\top} t_{i})$ and $\sigma$ is the sigmoid function. While BCE loss enhances the model's capability to differentiate between successes and failures in the training dataset, it risks fostering overconfidence in predictions due to the limited size of the robot dataset and insufficient feature granularity. This overconfidence may result in the model disproportionately favoring successful videos from the training set, while marginalizing others. Consequently, this approach could hinder the model's generalization capabilities in novel scenarios or tasks beyond the training dataset.

\textbf{Failure Prompts.} 
We aim to achieve a deeper understanding and identification of failure patterns and their root causes, rather than merely dismissing all unseen states. Our robot failure dataset comprises videos of failures, each accompanied by corresponding task description. By leveraging the distinct context of each failure, we seek to capture the unique precursors leading to each specific failure instance. Acknowledging the varied reasons for failures across different tasks, we propose the creation of a ``failure prompts pool'' to achieve this. This pool allows for flexible grouping and integration as input into the model, facilitating a nuanced and task-specific approach to understanding failures. For each task $\mathcal{T} \in \mathcal{D}^r_\mathcal{T}$, whose expression is $l_{\mathcal{T}}$, the task-specific prompt pool consists of $K$ learnable prompts:
$$
\mathbf{P^{f}_{\mathcal{T}}} = \{P_{\mathcal{T},1}^{f}, P_{\mathcal{T},2}^{f}, ..., P_{\mathcal{T},K}^{f}\} 
$$
where $P_{\mathcal{T},k}^{f} \in \mathbb{R}^{L_{p}^{f}\times \mathcal{D}} $ is a prompt with token length $L_{p}^{f}$ and the same embedding size $D$ as $y_\mathcal{T}$, where $y_{\mathcal{T}} = f_{e}^{t}(l_{\mathcal{T}})$ is the embedding features of $l_{\mathcal{T}}$.


We dynamically select suitable prompts for various videos depicting robotic task failures. For each task's failure videos, we utilize spherical K-means clustering to iteratively update the clustering centers at the end of each training epoch. This process enables the assignment of new pseudo-labels to the failure videos, effectively uncovering distinct failure themes specific to each task. Formally, for the task $\mathcal{T}$, we denote $\{v_i\}_{i=1}^{M_{\mathcal{T}}}$ as failure video features encoded by the vision encoder of the current epoch, where $M_{\mathcal{T}}$ devotes the number of videos in this task. 
The $\mathbf{i}$-th video's pseudo-label $\mathbf{q_{i}}\in\mathbb{R}^{K\times 1}$ and cluster centers $\mathbf{C}$ are obtained by minimizing the following problem:

\begin{equation}
\min _{\mathbf{C} \in \mathbb{R}^{d \times K}} \frac{1}{M_{\mathcal{T}}} \sum_{i=1}^{M_{\mathcal{T}}} \min _{\mathbf{q_{i}}}-v_i^{\top} \mathbf{C q_{i}}
\end{equation}


To ensure label stability, aligning clustering results across consecutive epochs is imperative. The assigned pseudo-labels are interpreted as indicators of the respective failure causes. We assume that these pseudo-labels, derived from clustering, succinctly encapsulate the semantic essence of each failed video, thereby elucidating the underlying reasons for failures. Consequently, we select failure prompts based on their corresponding pseudo-label $k$, leveraging this alignment to foster understanding of failure dynamics. For each task $\mathcal{T} \in \mathcal{D}^r_\mathcal{T}$, the input text embeddings of robot failure context are defined as follows:
$$
y^{f}_{\mathcal{T}, k} = [P^{f}_{\mathcal{T}, k}; y_{\mathcal{T}}]
$$
where ; denotes concatenation along the token length dimension and $k \in \{1...K\}$.
By learning $K$ failure prompts for each task, we aim for the model to identify $K$ failure causes per task.
Finally, for each epoch, we encode each failure context and get the feature set $t_{\mathcal{T}}^{f}=\{t_{\mathcal{T},1}^{f}, t_{\mathcal{T},2}^{f}, ..., t_{\mathcal{T},K}^{f}\}$.

In the previous section~\ref{sec:3.2}, semantically related video and text features are brought closer, and irrelevant or opposite semantic features are pushed away. Beyond that, we think a series of failure texts should also stay away from success videos. Hence, if we denote $(v_{i}, t_{i})$ as a video and language feature pair that completes task $\mathcal{T}$, its corresponding failure text features $t^{f}_{\mathcal{T}}$ should also be used as a negative sample. 
We modify video-language contrastive loss for each video-text pair $(v_{i}, t_{i})$ across human and robot domains as below:
\begin{equation}
\mathcal{L}_{VLC}^{v_{i}, t_{i}}=\frac{\exp \left(v^{\top}_{i} t_{i} / \tau\right)}{\sum_{j=1}^B \exp(v^{\top}_{i} t_{j} / \tau) + \sum_{k=1}^K\exp(v^{\top}_{i} t^{f}_{\mathcal{T},k} / \tau)}
\label{HH}
\end{equation}
Considering that different failure texts in the same failure text pool indicate distinct reasons for task failure, the failure video-text correspondences of different categories should be separated. Therefore, we define the failure's video-language contrastive loss for $(v^r_{i}, t^r_{i})$ from the failure robot sample $(x^r_{i}, l^r_{i}, r_i=0)$ as follows:
\begin{equation}
\mathcal{L}_{fVLC}^{v^r_{i}, t^r_{i}}=\frac{\exp({v^r_{i}}^{\top} t^{f}_{\mathcal{T},k^*}/\tau)}{\exp ({v^r_i}^{\top} t^r_i / \tau)+\sum_{k=1}^K \exp ({v^r_i}^{\top} t^{f}_{\mathcal{T},k} / \tau) }
\label{HH}
\end{equation}
where $k^*$ represents the index of the failure prompt (cluster) that the sample $(x^r_{i}, l^r_{i}, r_i=0)$ belongs to.

\section{Experiments}
\label{sec:experiments}
In our experiments, we aim to transfer video-language models with robust generalization into generalizable language-conditioned reward functions, utilizing robot video data from a minimal amount of tasks in a singular environment. We focus on studying how effectively our method Adapt2Reward can leverage a few successful and failed robot executions and to what extent doing so enables generalization to unseen environments and tasks and enhances the model’s ability to discriminate between successful and failed robot executions.
Concretely, we study the following questions:

1) Is Adapt2Reward able to generalize to new environments more effectively?

2) Is Adapt2Reward able to generalize to new tasks more effectively?

3) Can Adapt2Reward effectively improve the success rate of robots completing language-instructed tasks?

4) Can Adapt2Reward correctly assign rewards across different viewpoints or distracting objects?

\subsection{Experiment Setting}
\label{sec:4.1}
In our experiment, we employ the simulated environment from DVD~\cite{DVD}, adapted from Meta-World~\cite{MetaWorld}, built upon MuJoCo~\cite{MuJoCo} physical engine, featuring a Sawyer robot arm interacting with various objects including a drawer, faucet, cup, and a coffee machine. We utilize the video-language model Singularity~\cite{singularity}, pretrained on 17M human vision-language data and fine-tuned on the Something-Something-V2 (SS-V2) dataset~\cite{something-something-v2}. For human data in our training set, we utilize the Something-Something-V2 dataset, which includes 220,837 videos across 174 classes of basic human actions with diverse objects and scenes. We select videos from different human tasks. Our robot demonstration data, minimal in volume (560 videos for each task), encompasses successful and failed trajectories within the training environment. Following DVD~\cite{DVD}, we train the model with different tasks in different experiments, which will be detailed in the corresponding sections. For the hyper-parameters of Adapt2Reward, we adopt $B_h=B_r=B_f=8$ and $K=3$ as defaults. For convenience, we set the trade-off equal to 1 across all losses, including $\mathcal{L}_{CDC}$, $\mathcal{L}_{VLC}$, $\mathcal{L}_{fVLC}$.

\begin{figure}[t]
\centering
\subfloat[Environment Visualizations]{\includegraphics[width=0.45\textwidth]{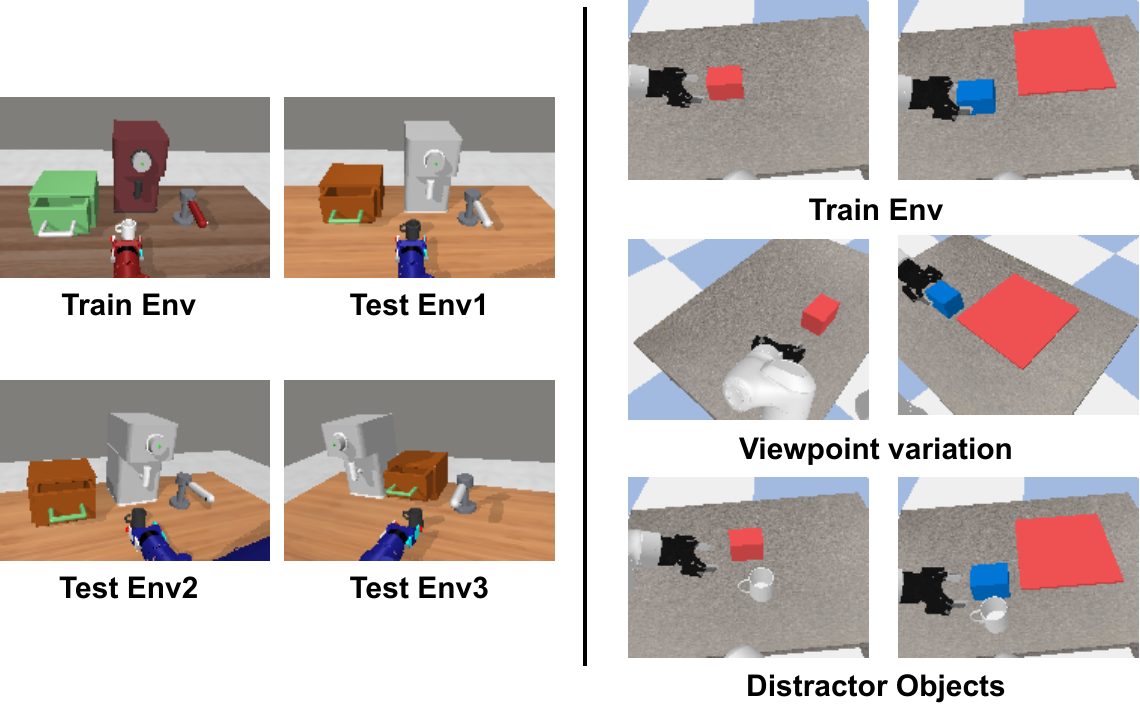}}
\hspace{2mm}
\subfloat[Environment Generalization Results]{\includegraphics[width=0.44\textwidth]{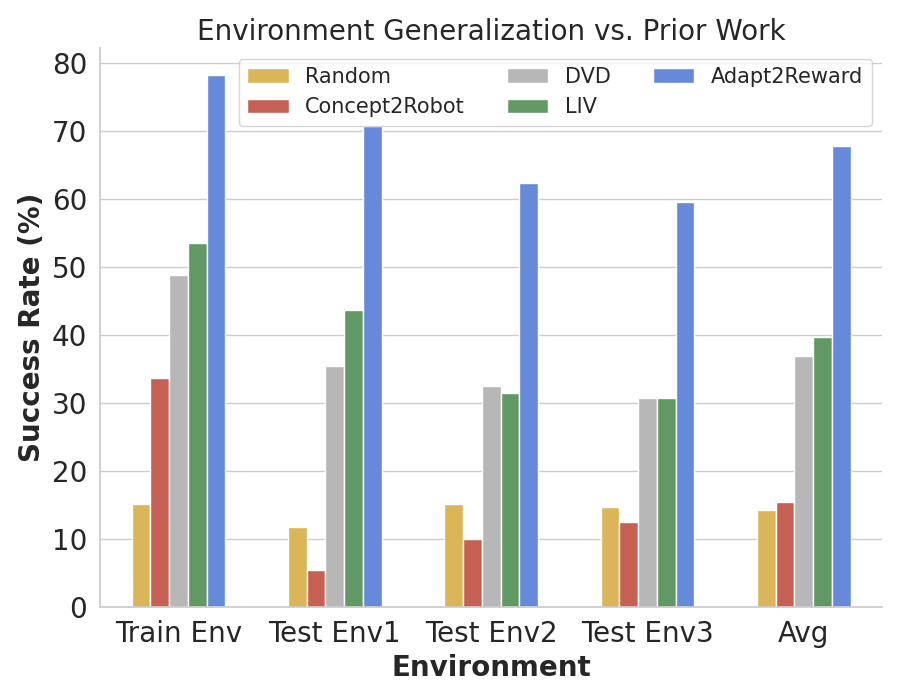}}


\caption{(a) MetaWorld environments (left) consist of the original training environment and three test environments with color, viewpoint, and object arrangement modifications. Concept2Robot environments (right) include the training environment and the testing environments that change the viewpoint or include distractor objects.
(b) 
Comparison of Random Policy, VMPC with Concept2Robot, DVD, LIV, and Adapt2Reward in MetaWorld environments. 
The depicted bars represent the mean success rate across 4 target tasks computed over 3 seeds of 100 trials.
}
\label{fig_env}
\end{figure}

\textbf{Task Execution.}
Once the reward function $\mathcal{R}{\theta}$ is trained, we use it to select actions with visual model predictive control (VMPC)~\cite{EmbedTC, VisualFM, LearningLD}, which uses a learned visual dynamics model to plan a sequence of actions. We condition $\mathcal{R}{\theta}$ on a language instruction $l$ for the target task $\mathcal{T}$, using predicted similarity as the reward.
We train an action-conditioned video prediction model $ p_{\phi}(s_{t+1:t+H}|s_{t}, a_{t:t+H})$ with the FitVid framework~\cite{FitVid}. Given an input image $s_{t}$, we sample G action trajectories of length H and use $p_{\phi}$ to predict their future trajectories $\left\{s_{t+1:t+H}\right\}^{g}$. We then score the similarity between the task instruction $l$ and each predicted trajectory using $\mathcal{R}_{\theta}$. The action trajectory with the highest score is executed.

\begin{table*}[t]
\centering
\scalebox{0.8}{
\begin{tabular}{ccccccc}
\toprule
 & \makecell{Human Only}  &\makecell{Robot Only}  &\makecell{Robot +\\3Human Tasks}  &  \makecell{Robot +\\6Human Tasks}  & \makecell{Robot +\\9Human Tasks}  & Avg         \\
\midrule
Random & 15.25(0.31) & - & - & - & - & 15.25(0.31) \\
Concept2Robot & 33.75(0.20) & -  & - & - & - & 33.75(0.20)\\
DVD & - & 19.00(1.07)  & 28.25(0.20) & 31.92(2.01) & 40.08(0.31) & 29.81(0.21) \\
LIV & - & 27.58(1.65)& 36.33(0.42)& 38.83(1.03)& 39.00(1.24)& 35.44(0.51)\\
Adapt2Reward & -  & \textbf{29.92(0.83)}  & \textbf{44.67(0.77)} & \textbf{69.33(2.42)} & \textbf{67.00(1.14)} & \textbf{52.73(0.64)} \\
\bottomrule
\end{tabular}}
\caption{\textbf{Task Generalization with Different Reward Models.} Compared with Random Policy, Concept2Robot, DVD, and LIV, Adapt2Reward achieves significantly enhanced performance in novel tasks.  
We report the average success rate on 4 target tasks, computed over 3 seeds of 100 trials, with the standard deviation in parentheses.}
\label{task generalization table}
\end{table*}

\subsection{Environment Generalization}
\label{Environment Generalization}
Ideally, a reward function should be robust to naturalistic visual variations, including changes in camera angles, object colors, or arrangements. Given the impracticality of re-annotating and retraining for each new condition, it's crucial to develop a visually resilient reward function.
To explore environmental generalization, we use four progressively challenging variants of a given environment, as depicted in Fig.~\ref{fig_env} (b). These consist of the original training environment (Train Env) with task demonstrations, and three test environments with modifications: Test Env 1 (changed colors), Test Env 2 (changed colors and viewpoint), and Test Env 3 (changed colors, viewpoint, and object arrangement). We evaluate our method on four simulated tasks: (1) \textsl{closing drawer}, (2) \textsl{moving cup away from the camera}, (3) \textsl{moving the handle of the faucet}, and (4) \textsl{pushing cup from left to right}. 
To evaluate our method, we train it on robot videos from the training environment, covering the four tasks with both successful and failed executions. We also incorporated human data from the SS-V2 dataset, encompassing these four tasks and three additional tasks for training.

We compare our approach with previous works. Concept2Robot~\cite{Concept2Robot}, which uses a pre-trained 174-way action classifier from the SS-V2 dataset, taking the classification score of the predicted robot video as the reward. Unlike Concept2Robot conditioning on the category of the task, DVD~\cite{DVD} calculates rewards by training a domain-agnostic video discriminator to measure the similarity between robot videos and human demonstrations. DVD also involved varying amounts of human data from the SS-V2 dataset, but not including failed robot executions. 
LIV~\cite{liv} generates cross-modal embeddings with temporal coherence and semantic alignment, calculating rewards based on current state images and language objectives. We fine-tune LIV and train DVD on the same successful human and robot task videos as Adapt2Reward.
We also include a comparison to a random policy. 
For all reward functions, we use the aforementioned VMPC for action selection to ensure a fair comparison.
In Fig.~\ref{fig_env}, we compare Adapt2Reward with 7 human tasks to these prior methods. Across all environments, Adapt2Reward significantly outperforms the three comparison methods on target tasks, exceeding the best-performing method by over 28\% on average.

\subsection{Task Generalization}
In this experiment, we investigate the impact of incorporating human data on the reward function's capacity to generalize across new tasks. Following DVD~\cite{DVD}, we train the Adapt2Reward model on robot videos from three distinct tasks within the training environment without utilizing any robot data from the target tasks. Specifically, these three tasks are used for training: (1) \textsl{opening drawer}, (2) \textsl{pushing cup from right to left}, (3) \textsl{poking cup so lightly that it doesn't or almost doesn't move}; the four target tasks are listed in the Section~\ref{Environment Generalization}. Additionally, we incorporate a variable amount of human data into our training regimen. As shown in Table~\ref{task generalization table}, we compare our method with the methods described in our prior experiment. Given our focus on task generalization, all evaluations are conducted within the training environment. To further evaluate the influence of human data on enhancing generalization capabilities, we conduct comparative analyses between our method and prior approaches across various human tasks in Table~\ref{task generalization table}.
Following DVD, `Robot Only' means the reward model is trained solely with robotic data from three training tasks, without any data from four target tasks or any human data. `Robot + 3 Human Tasks' refers to training with both robot and human videos from the three training tasks. `Robot + 6 Human Tasks' further incorporates human data from three additional tasks (different from the training and target tasks).
In Table~\ref{task generalization table}, the success rates of Adapt2Reward on novel tasks surpass those of previous methods by nearly 30\%. 
Our analysis also shows that training with human videos enhances task generalization by approximately 15-40\% over Robot Only, likely due to mitigating overfitting and maintaining human data-derived knowledge. Incorporating 6 human tasks significantly boosts model performance, affirming the efficacy of a hybrid training approach in bridging human and robotic domain knowledge.
\begin{figure}[t]
\centering
\subfloat[Failure data]{\includegraphics[width=0.24\textwidth]{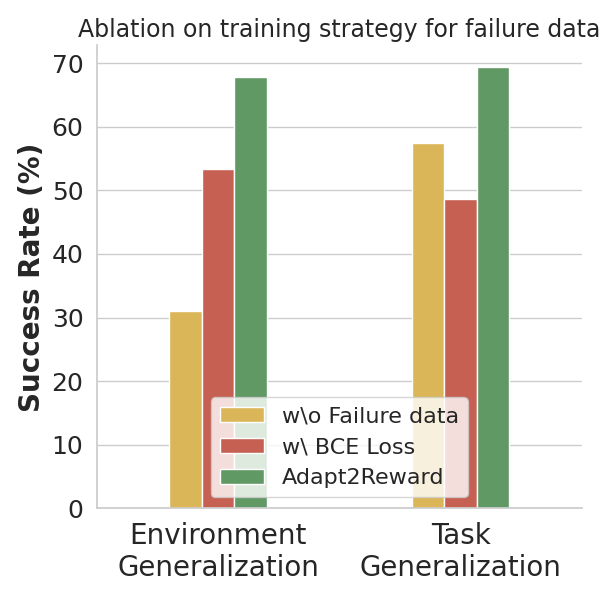}}
\subfloat[Cluster Number K]{\includegraphics[width=0.24\textwidth]
{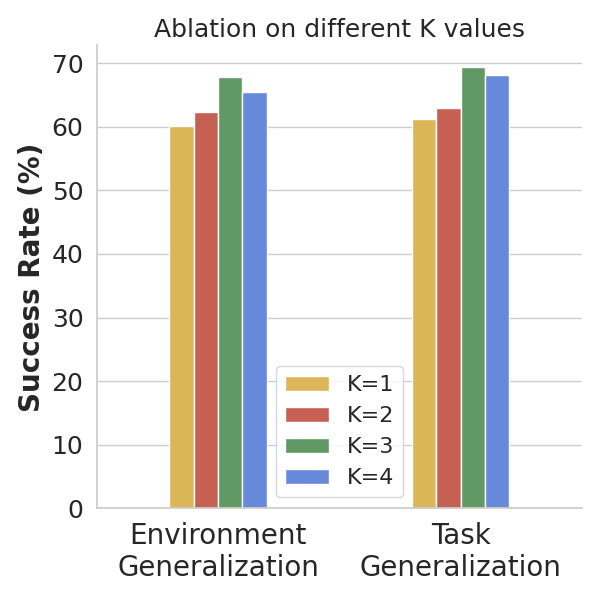}}
\subfloat[Failure Data Source]{\includegraphics[width=0.24\textwidth]
{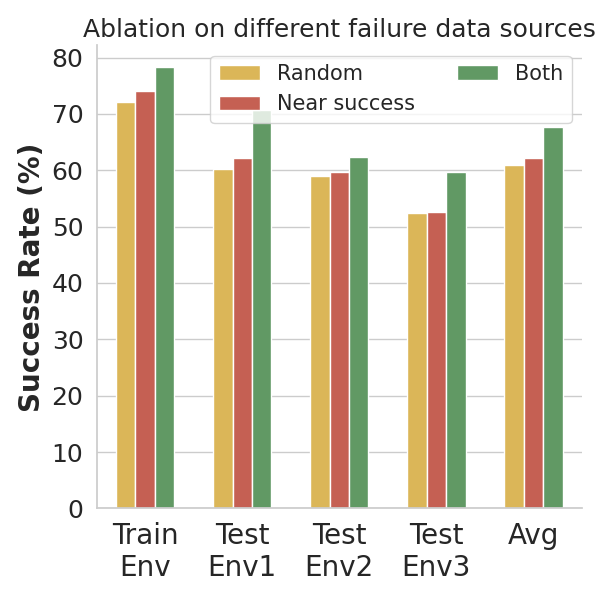}}
\subfloat[Distribution]{\includegraphics[width=0.24\textwidth]
{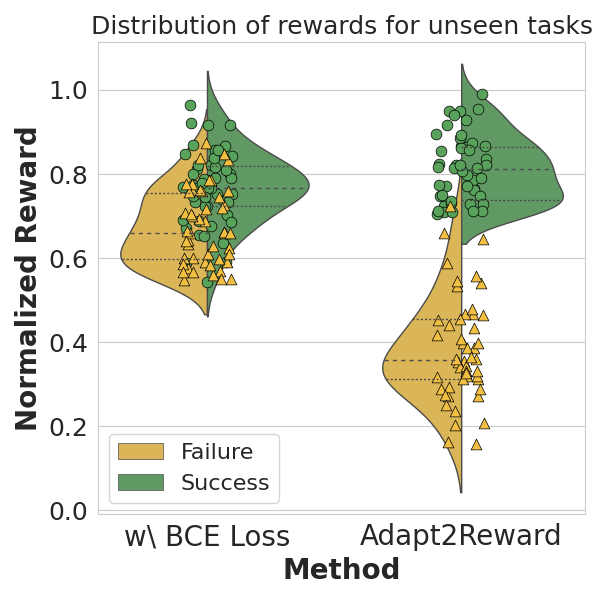}}
\caption{\textbf{Ablation study.} (a) Different training methods for failure data. (b) Varying K. (c) Training with different sources of failure data. (d) The distribution differences in rewards obtained by different reward methods for an unseen task, with scattered points representing normalized reward values for different trajectories.}
\label{fig:ablation study}
\end{figure}

\subsection{Ablation Study}

In this study, we aim to assess the influence of failure data on Adapt2Reward, contrasting two training methodologies: one devoid of failure data and the other utilizing Binary Cross-Entropy (BCE) loss for distinguishing successful from unsuccessful videos. In Fig.~\ref{fig:ablation study}(a), the average success rates in environmental and task generalization experiments are presented. The results indicate a substantial enhancement in the model's generalization ability with the inclusion of failure data, likely due to learning from a broader range of characteristics and patterns. Moreover, leveraging failure prompts for understanding the causes and contexts of failures proved more effective for generalization than merely using BCE loss, which may compromise the model's learned knowledge from human data. 
Furthermore, in Fig.~\ref{fig:ablation study}(b), our ablation study on the cluster number $K$, indicating the number of failure categories per task, shows our approach is not very sensitive to the hyper-parameter and $K=3$ as the optimal value.

We also assess how different sources of failure data affect generalization. We use two methods to collect failure data for training: (1) random exploration and (2) near-success scenarios (achieved by adding noise to successful trajectories or collecting from pre-trained RL models). In Fig.~\ref{fig:ablation study}(c), we present success rates of Adapt2Reward with failure data from different source across several environments. Results indicate that combining failure data from varied sources outperforms using data from a single source, possibly due to a wider range of failure patterns captured from diverse origins.

To demonstrate the superior generalization of Adapt2Reward over models trained with BCE loss for unseen tasks, Fig.~\ref{fig:ablation study}(d) showcases the reward distribution for various trajectories. Models using BCE loss incorrectly assigne high rewards to failed trajectories. In contrast, Adapt2Reward accurately distribute rewards, clearly differentiating between successful and failed trajectories.


\begin{figure*}[t]
\centering
\includegraphics[scale=0.3]{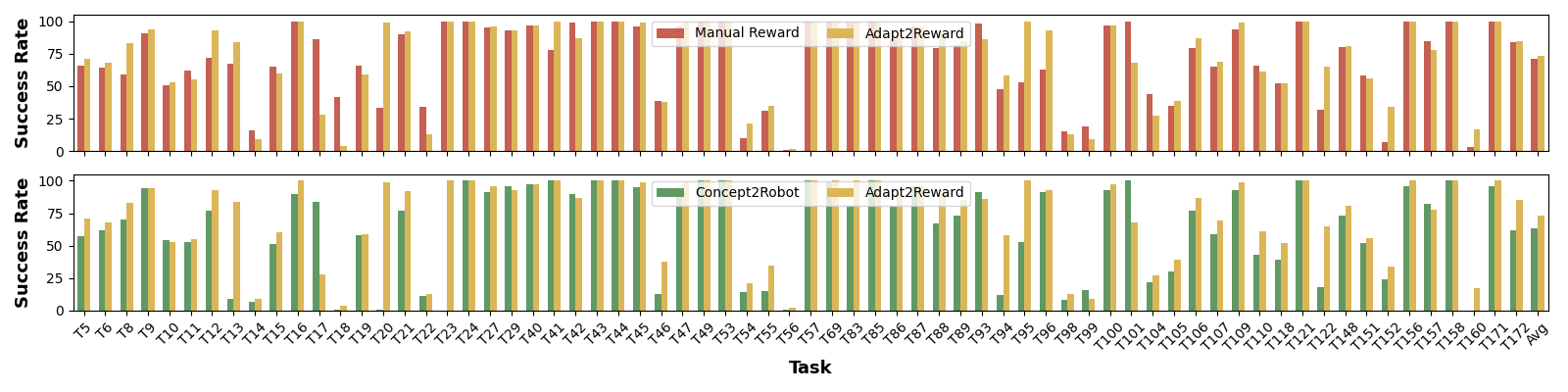}
\caption{\textbf{Task Generalization in C2R-Envs.} We report the success rates of reinforcement learning on 68 tasks with manually crafted rewards, Concept2Robot, and Adapt2Reward. The majority of policies obtained by Adapt2Reward match or exceed ``manually crafted reward''. In comparison to Concept2Robot, Adapt2Reward demonstrated superior performance across most tasks.}
\label{fig_exp4}
\end{figure*}

\subsection{Concept2Robot Environments Efficacy}

To further evaluate our approach, we use Pybullet to simulate each environment associated with 68 tasks. Similar to Concept2Robot~\cite{Concept2Robot} Environments (C2R-Envs), the robot is a simulated 7-DoF Franka Panda robot arm with a two-fingered Robotiq 2F-85 gripper. We consider a similar setup as described in Section~\ref{sec:4.1}, where Adapt2Reward is now trained on robot videos of the 12 robot tasks from the simulation environment including success and failed execution (500 videos for each task), as well as human data of all tasks in SS-V2, and measure performance across seen and unseen robot tasks. In this experiment, unlike the VMPC approach for selecting actions in the previous experiments, we follow the method used in the Concept2Robot~\cite{Concept2Robot} and leverage the open-loop trajectory generator, which combines deep Deterministic Policy Gradients (DDPG) \cite{DDPG} with the Cross-Entropy Method (CEM)~\cite{CEM} to solve the problem.

\begin{table}[t]
    \begin{subtable}{.5\linewidth}
    \centering
    \resizebox{!}{0.82cm}{
        \begin{tabular}{ccc}
        \toprule
        Task                           & Concept2Robot & Adapt2Reward \\
        \midrule
        pushing box from left to right & 86(-5)          & \textbf{91(+5)}         \\
        pushing box onto book          & 54(-37)          & \textbf{81(-12)}         \\
        moving book up                  & 89(-6)          & \textbf{97(-2)}         \\
        moving book down               & 95(-5)          & \textbf{100(0)}        \\
        \bottomrule
        \end{tabular}
    }
    \caption{Viewpoint variation.}
    \end{subtable}%
    \begin{subtable}{.5\linewidth}
    \centering
    \resizebox{!}{0.82cm}{
        \begin{tabular}{ccc}
        \toprule
        Task & Concept2Robot & Adapt2Reward \\
        \midrule
        pushing box from left to right & 37(-54) & \textbf{66(-20)} \\
        pushing box onto book & 39(-52) & \textbf{61(-32)} \\
        moving box away from box & 53(-44) & \textbf{99(+2)} \\
        moving box closer to box & 71(-19) & \textbf{91(+4)} \\
        \bottomrule
        \end{tabular}
    }
    \caption{Distractor Objects.}
    \end{subtable}
\caption{\textbf{Visual Robustness.} We report the success rates of Concept2Robot and Adapt2Reward trained with different viewpoints or distractor objects in C2R-Envs. Parentheses show the percentage change from an unchanged viewpoint.} 
\label{Concept2Robot_robust}
\end{table}
In Fig.~\ref{fig_exp4}, we display the success rate of 68 tasks under different reward functions. 
We find that Adapt2Reward-based strategies often equal or surpass ``manually crafted reward'' strategies. This advantage likely stems from Adapt2Reward's provision of dense reward scores, which favorably assess near-success attempts in contrast to binary ground truth rewards.
In further comparisons, Adapt2Reward consistently outperform Concept2Robot, which utilizes the classifier as a reward function, across several tasks, notably Task12 (\textsl{dropping a box onto a book}) and Task95 (\textsl{pushing a box off a book}). This superiority may result from Concept2Robot's generic templates representing objects as ``[Something]'' during training, which compromises task-specific object interaction recognition, hindering accurate success state and object relation identification.

Also, We examine reward function robustness by varying viewpoints and objects. Although Adapt2Reward was trained on frames from a fixed camera perspective, we evaluated its performance under different camera angles. As shown in Table~\ref{Concept2Robot_robust} (a), changing camera angles significantly impacted Concept2Robot (success rates dropped by 5\%-37\%), whereas Adapt2Reward showed greater stability (success rates decreased by less than 12\%).
As shown in Table~\ref{Concept2Robot_robust} (b), Adapt2Reward outperformed Concept2Robot by 20-46\% in success rates under this new visual arrangement, demonstrating its robustness to visual variations.

\section{Conclusion}
\label{sec:conclusion}
In our work, we underscore the importance of integrating failed robotic videos into reward learning alongside human videos, addressing the disparity between human video-language datasets and robotic datasets in this context. We introduce learnable failure prompts as an effective method to capture patterns of robotic failures, significantly augmenting the model's adaptability and applicability. Our method shows superior generalization in robotic environments, i.e., Meta-World and Concept2Robot environments, where our reward models are effectively adapted to unseen environments and instructions.

\section*{Acknowledgment}
This work was supported in part by The National Nature Science Foundation of China (Grant No: 62273303, 62303406), in part by the Key R\&D Program of Zhejiang Province, China (2023C01135), in part by Ningbo Key R\&D Program (No.2023Z231, 2023Z229), in part by Yongjiang Talent Introduction Programme (Grant No: 2022A-240-G, 2023A-194-G).

{
    \small
    \bibliographystyle{splncs04}
    \bibliography{main}
}

\end{document}


\title{Adapt2Reward: Adapting Video-Language Models to Generalizable\\Robotic Rewards via Failure Prompts} 

\titlerunning{Adapt2Reward}

\author{Yanting Yang \inst{1}\thanks{Equal contribution;$^{\dagger}$ Corresponding author}  \and 
Minghao Chen\inst{2*\dagger}\and
Qibo Qiu\inst{1,3}\and
Jiahao Wu\inst{4} \and
Wenxiao Wang\inst{1} \and
Binbin Lin\inst{1} \and
Ziyu Guan\inst{5} \and
Xiaofei He\inst{6}
}

\authorrunning{Y.~Yang et al.}

\institute{School of Software Technology, Zhejiang University \and
School of Computer Sciene and Technology, Hangzhou Dianzi University \and
China Mobile (Zhejiang) Research \& Innovation Institute
 \and The Hong Kong Polytechnic University \and
 School of Computer Sciene and Technology, Xidian University
 \and State Key Lab of CAD\&CG, Zhejiang University
}


\clearpage
\setcounter{page}{1}
\appendix

\section{Training Details}
\textbf{Dataset Details.} In prior work~\cite{singularity}, the action recognition dataset Something-Something v2 (SSv2)~\cite{something-something-v2} was transformed into a video and language dataset. In our work, we employ 174 templates from SSv2 (e.g., ``Opening [Something]") as action descriptions for the videos, and utilize annotated labels from SSv2 (e.g., ``Opening drawer") as task descriptors for the videos. In the experiments of the first three chapters, we select videos from the following 12 different human tasks which represent different actions, where each task has 853-3170 training videos: 1) Closing [Something], (2) Moving [Something] away from the camera, (3) Moving [Something] towards the camera, (4) Opening [Something], (5) Pushing [Something] from left to right, (6) Pushing [Something] from right to left, (7) Poking [Something] so lightly that it doesn't or almost doesn't move, (8) Moving [Something] down, (9) Moving [Something] up, (10) Pulling [Something] from left to right, (11) Pulling [Something] from right to left, (12) Pushing [Something] with [Something]. As delineated in DVD~\cite{DVD}, the tasks chosen are apt for single-arm setups in desktop scenarios, covering diverse actions. The selection of the initial seven tasks aligns with their pertinence to simulated environments, while the inclusion of the remaining tasks, not guided by specific criteria, allows for their potential replacement with an alternate set of eight suitable tasks. Similarly, in the experiment of Section 4.5, we select videos from the 68 different human tasks related to the Concept2Robot simulation environment. In our experimentation, for each specific task, Adapt2Reward is trained using all corresponding human videos from the Something-Something V2 dataset. To enhance the model's capacity for detailed object recognition, we incorporate task descriptions with annotated labels, coupled with the videos, as model inputs.
\begin{figure}[]
\centering
\includegraphics[scale=0.29]{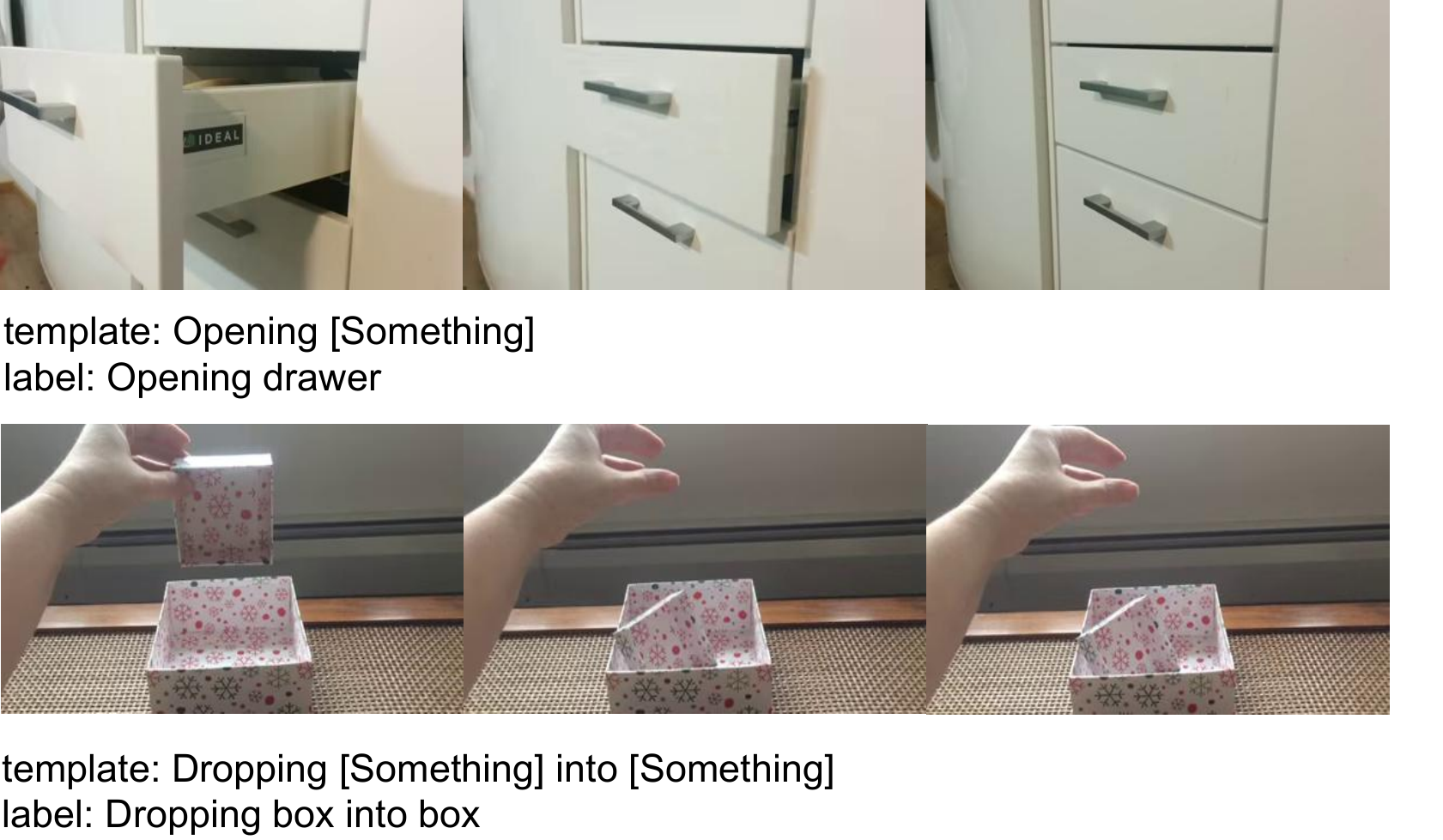}
\caption{\textbf{Two samples in Something-Something v2.}
}
\label{suppl_ssv2_example}
\end{figure}

In the Meta-World simulation environment, our Adapt2Reward is trained with 280 successful and 280 failed robotic videos per task. 
The successful robot data consists of two parts: the first part is collected by generating random action sequences and filtering them using ground truth rewards, and the second part is gathered by a hard-coded script with uniform noise from $[-0.03, 0.03]$. It performs flawlessly successfully by manipulating a robot arm. We posited that the failure data should encompass both failed attempts during random exploration and those nearly successful to reflect realistic learning scenarios. Therefore, the failure robotic data are partly gathered by random shooting and a ground truth reward to each task and partly collected by Soft Actor-Critic (SAC)~\cite{SAC} trained on these ground truth rewards, capturing a comprehensive spectrum of failure modes. For the Concept2Robot simulation environment, in addition to varying amounts of human videos, Adapt2Reward is trained on 500 robot success videos and 500 robot failure videos of 12 tasks. These videos are collected by training single-task policies with a hand-designed, task-specific, binary reward signal. In particular, we also collect failure data during random exploration.

\textbf{Training Settings.}
We use the video-language model Singularity~\cite{singularity}, pretrained on 17M human vision-language datasets, and fine-tune it on the Something-Something-V2 (SS-V2) dataset~\cite{something-something-v2}. To recognize fine-grained temporal relationships between video frames, we utilize its multi-frame variant, Singularity-Temporal. This variant introduces a two-layer temporal transformer encoder after the vision encoder, feeding its outputs into the multi-modal encoder. In our experiments, we freeze the text encoder and fine-tune the other modules on mixed human-robot data. We use a learning rate of $5e-6$ with AdamW, momentum values of $0.9$ and $0.999$, and a weight decay of $0.02$. Notably, we employ a higher learning rate of $0.001$ for prompts, as they are randomly initialized. Our Adapt2Reward model is trained for 10 epochs using a cosine decay schedule. We set our batch sizes $B_h=B_r=B_f=8$ and $K=3$ as default parameters. We maintain a uniform trade-off of 1 across all losses, including $\mathcal{L}_{CDC}$, $\mathcal{L}_{VLC}$, and $\mathcal{L}_{fVLC}$. Additionally, we use 4 frames for both training and inference, and apply random resize, crop, and color jitter as augmentation techniques across all videos.

%
\begin{figure*}[t]
\centering
\includegraphics[scale=0.3]{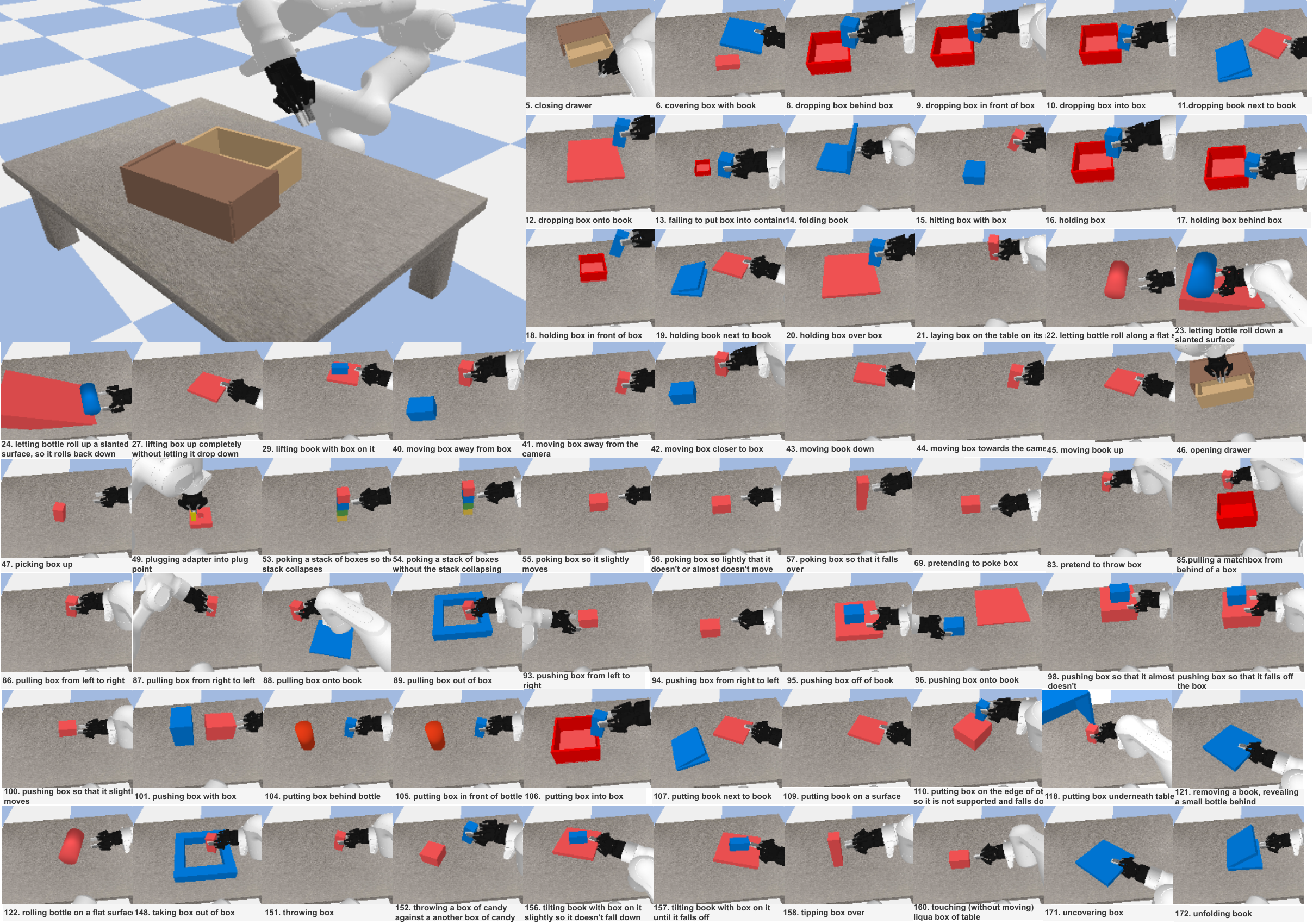}

\caption{\textbf{The modified Concept2Robot environment in our experiments} }
\label{suppl_concept2robot_env}
\end{figure*}

\textbf{Training FitVid.} 
To evaluate Adapt2Reward’s performance on
potentially unseen tasks with a given language instruction, we employ visual model predictive control with FitVid~\cite{FitVid} visual prediction models trained on datasets autonomously collected
in each environment. FitVid introduces a convolutional non-hierarchical variational model with a fixed prior that can significantly overfit the video prediction datasets. We use the same architecture and default hyperparameters as the original paper. 

However, long-distance predictions remain a major cause of low-quality predictions by models. In our experiments, we do not need images of every frame to calculate rewards. Therefore, unlike FitVid, which predicts the next frame of the video frame by frame, we splice 4 frames of actions and feed them to the model together with the current state image, and the model will output the state image after executing the 4 actions. Therefore, a task of predicting 60 frames frame by frame based on an initial state of one frame will be transformed into a task of predicting 16 frames from the initial state. 
For each of the four simulation environments in which we evaluate Adapt2Reward, we collect 10,000 random episodes, each with 60 total frames, of the agent interacting in that environment and train FitVid for 200,000 epochs on all of the data. The models are trained to predict the next 12 frames given an input of init frames. In the inference, the model predicts the next 16 frames given an input of init frames.

\section{Experimental Details}
\textbf{Meta-World simulation Environments.} 
We utilize the same MuJoCo simulation based on the Meta-World~\cite{MetaWorld} environments as used in DVD~\cite{DVD}. In this simulation, the state space comprises RGB image observations with dimensions $[180, 120, 3]$. Our action space includes continuous actions for the robot’s gripper’s linear and angular velocities, and discrete actions for the gripper's open/close movement, resulting in a total of five dimensions.

\textbf{Concept2Robot simulation Environments.}
Following the environments in Concept2Robot~\cite{Concept2Robot}, we selected 68 out of the 174 tasks that fit our environment setting. We excluded 10 problematic environments that either couldn't be executed or had irrelevant actions and commands. Additionally, we modify the environment and manually craft reward functions for some tasks to ensure their actions and verbal commands are relevant and unaffected by extreme object states. The remaining 96 tasks were unsuitable as they either required dual-arm manipulation or were challenging to simulate in PyBullet, such as “tearing something”. For the selected tasks, we used the same task IDs as in SSv2, detailed in Figure~\ref{suppl_concept2robot_env}. Unlike Concept2Robot~\cite{Concept2Robot}, we used labels as textual descriptions, for example, “dropping box into box”, instead of templates. We simulated each of the 68 tasks in PyBullet. Our robot model is a 7-DoF Franka Panda robot arm equipped with a two-fingered Robotiq 2F-85 gripper. A static camera captures the environment state as RGB images, downscaled to 120×160. At each reset after an RL episode, the manipulation objects in each environment are initialized with a random pose within manually set bounds for each task.

\textbf{Details of Evaluation.}
In the previous study, DVD, trajectories of sixty frames were divided into three twenty-frame segments for action planning. Each twenty-frame segment had 100 action sequences uniformly and randomly sampled. The trajectory with the highest similarity to human demonstrations, as computed by the DVD algorithm, was chosen for environmental execution. The initial state for each subsequent segment was the final state from the previous trajectory. We contend that this segmented approach might lead to longer inference times, particularly with video prediction models that have a high number of parameters. We propose directly comparing the full sixty-frame video with human demonstrations, arguing that it is more intuitive than comparing segmented twenty-frame parts. Segment planning leads to the robot's task performance video being divided into three parts, with each part's similarity to the human demo calculated separately. This could yield inaccurate similarity scores, as DVD trains on unsegmented robot and human videos.

In our experiment, we use continuous sixty-frame trajectories for uninterrupted planning. We randomly generate 300 action sequences and employ the FitVid model to predict 300 six-second videos from initial images. Based on the task's language command, we calculate the rewards for these videos and select the highest-scoring trajectory for execution. For DVD, we execute the trajectory most similar to human demonstrations. 
For Concept2Robot, we execute the trajectory with the highest classification score in the environment.

We evaluate the following four target tasks: 1) Closing drawer, which is defined as the last frame in the 60-frame trajectory having the drawer pushed in to be less than 0.05, where it starts open at 0.07; 2) Moving cup away from the camera, move the cup away from the camera at least 0.1 distance compared to the initial distance; 3) Moving the handle of the faucet,  moving the handle more than 0.01 distance, where it starts at 0; 4) Pushing cup from left to right,  pushing the cup at least 0.05 distance from left to right. We run 100 trials for 3 different seeds for each task for every method in all experiments.

In the Concept2Robot environments, following the Concept2Robot~\cite{Concept2Robot}, we combine Deep Deterministic Policy Gradients (DDPG) ~\cite{DDPG} with the Cross Entropy Method (CEM)\cite{CEM} to improve action selection. The DDPG critic network approximates the Q-value function, and the actor network's output initiates the CEM. CEM then searches for a better action near the initial one, aiming to optimize decision-making\cite{QTOptSD}. Similar to Concept2Robot\cite{Concept2Robot}, we use task-specific success metrics to evaluate the tasks. We report the success rate for each task in 100 trials for every method across all experiments.

\begin{table}[t]
\centering
\scalebox{0.85}{
\begin{tabular}{ccc}
\toprule
Failure Prompt Length & Environment Generalization & Task Generalization \\
\midrule
1  & 64.58 & 60.75 \\
2  & \textbf{67.76} & \textbf{69.33} \\
4  & 66.65 & 63.25 \\
\bottomrule
\end{tabular}}
\vspace{4mm}
\caption{\textbf{Ablation study on Failure Prompt Length.} We report the average success rate of Adapt2Reward in environment generalization and task generalization experiments using different failure prompt lengths.}
\label{ablation on failure prompt length}
\end{table}

\begin{table}[t]
\centering
\scalebox{0.85}{
\begin{tabular}{ccc}
\toprule
Task                           & Concept2Robot & Adapt2Reward \\
\midrule
pushing cup from left to right &   88        &    \textbf{99}      \\
pushing cup onto book          &   18        &     \textbf{70}     \\
moving box up                  &   98       &     \textbf{100}     \\
moving box down                &   84        &      \textbf{100}   \\
\bottomrule
\end{tabular}}
\vspace{4mm}
\caption{\textbf{Change Objects.} We report the success rates trained Concept2Robot and Adapt2Reward. Adapt2Reward outperformed Concept2Robot, achieving a success rate higher by 2-52\%}
\label{change_object}
\end{table}

\begin{figure}[t]
\centering
\subfloat[Vision Encoder]{
\includegraphics[scale=0.25]{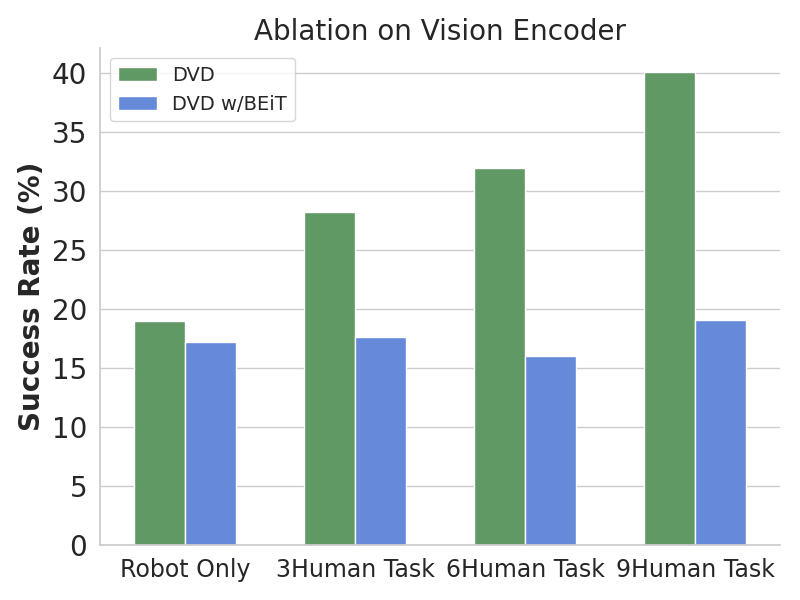}}
\subfloat[Different number of robot data]{
\includegraphics[scale=0.25]{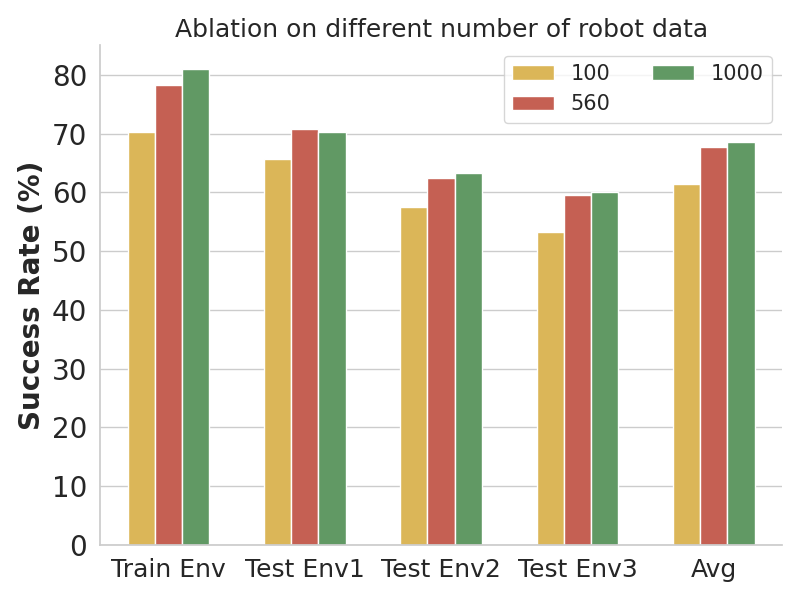}
}

\subfloat[Clustering method]{
\includegraphics[scale=0.25]{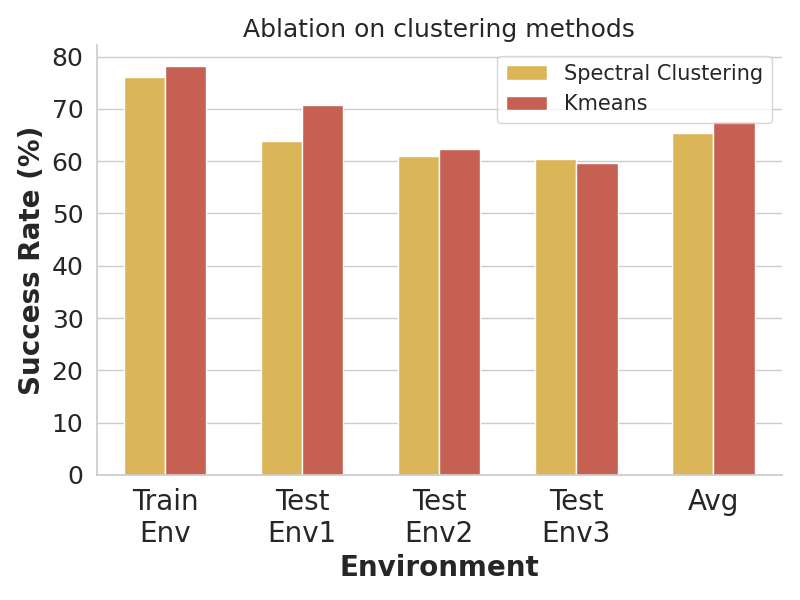}
}
\subfloat[t-SNE visualization]{
\includegraphics[scale=0.3]{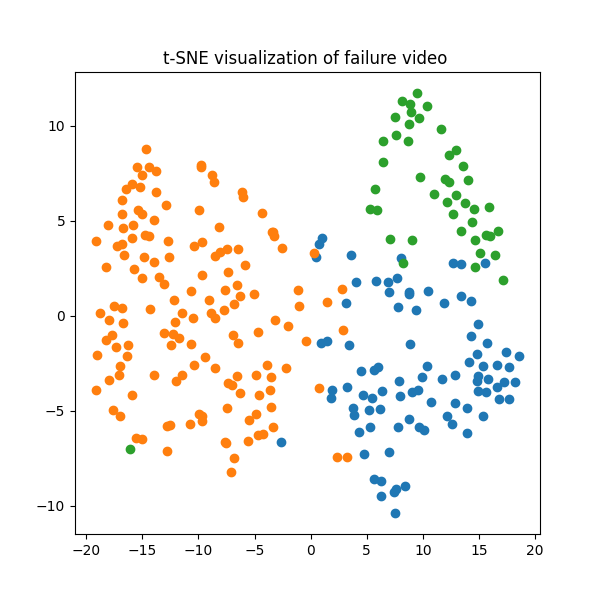}
}
\caption{(a) We present the average success rate of DVD in task generalization experiments across different vision encoders. (b) We show the average success rate of Adapt2Reward in environment generalization experiments with different amounts of training robot data, where 560 is the number adopted in the main text. (c) We display the average success rate of Adapt2Reward in four environments with different clustering methods. (d) We use t-SNE to visualize the failed videos from the "moving the handle of a faucet" task, with different colors representing different failure prompts.
}
\label{fig_suppl_dvd_backbone}
\end{figure}

\begin{figure*}[t]
\centering
\includegraphics[scale=0.42]{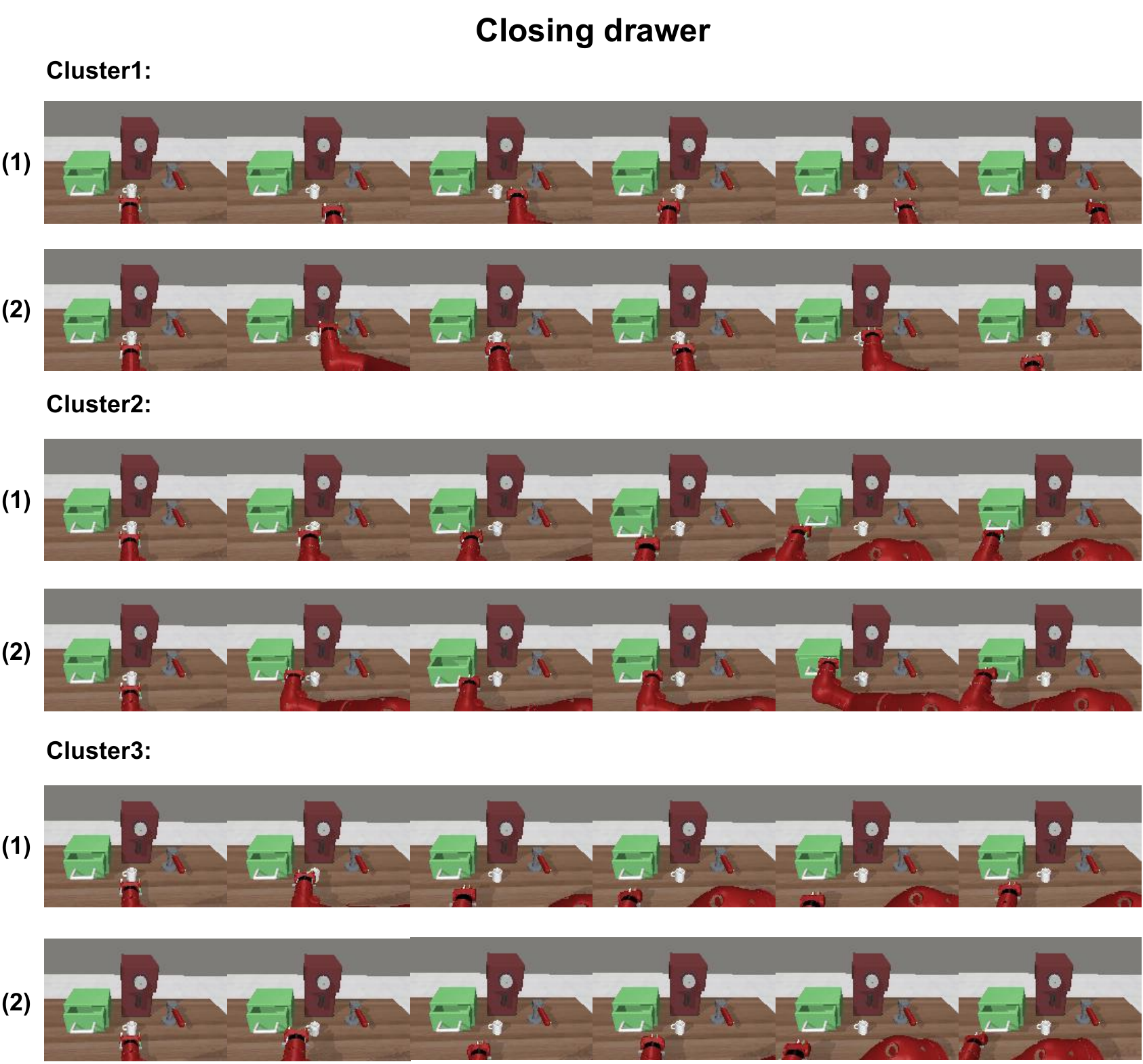}
\caption{\textbf{Examples of Clustering for the Task of Closing Drawer.} We display failure videos from the same cluster for the task of closing drawer. With $K$ set to 3, we provide two examples per cluster.}
\label{suppl_task5_cluster_demos}
\end{figure*}
\begin{figure*}[t]
\centering
\includegraphics[scale=0.42]{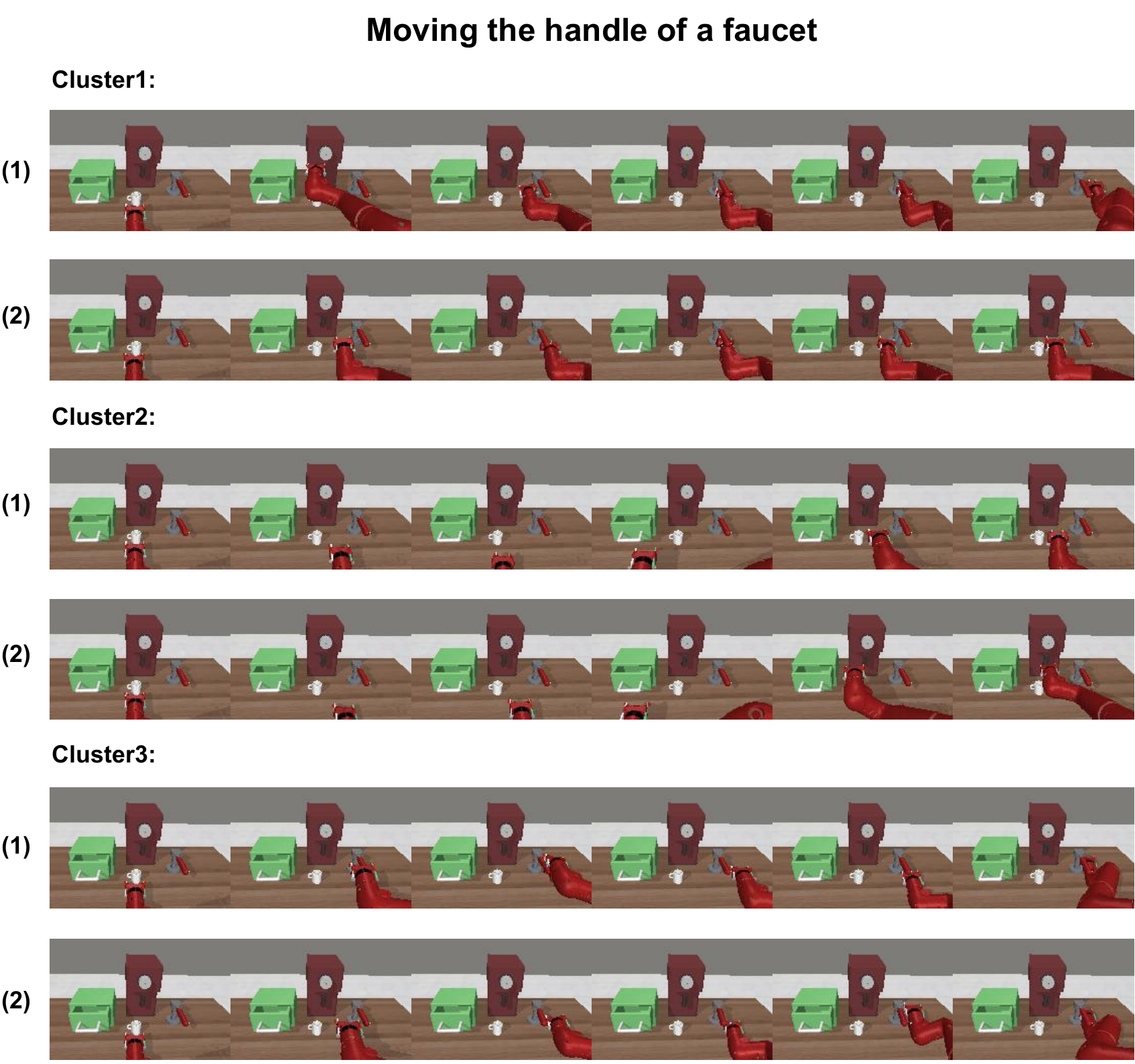}

\caption{\textbf{Examples of Clustering for the Task of Moving the handle of a faucet.} We display failure videos from the same cluster for the task of moving the handle of a faucet. With $K$ set to 3, we provide two examples per cluster.}
\label{suppl_task33_cluster_demos}
\end{figure*}

\begin{figure}[]
\centering
\includegraphics[scale=0.3]{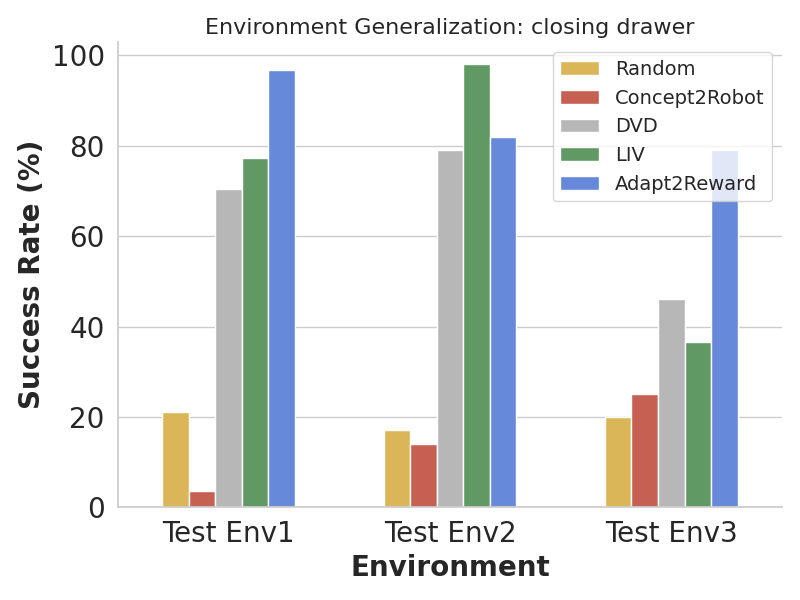}
\includegraphics[scale=0.3]{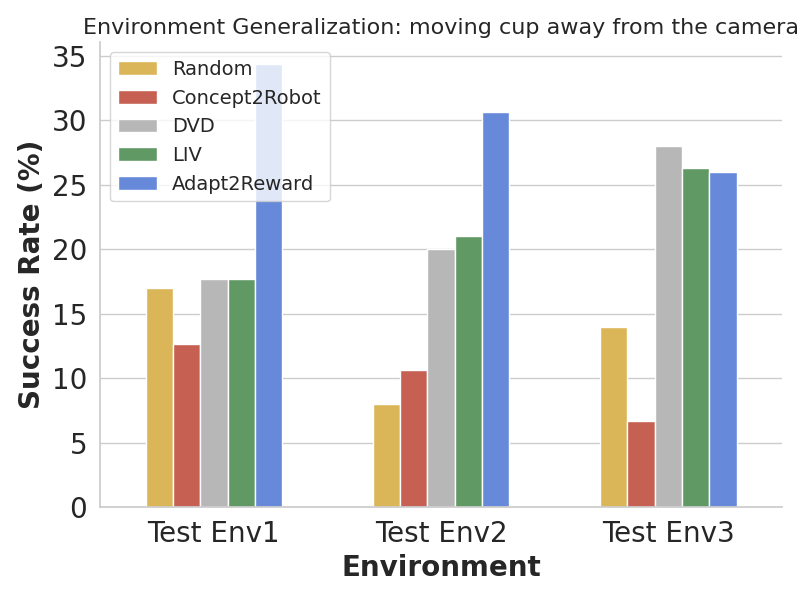}
\includegraphics[scale=0.3]{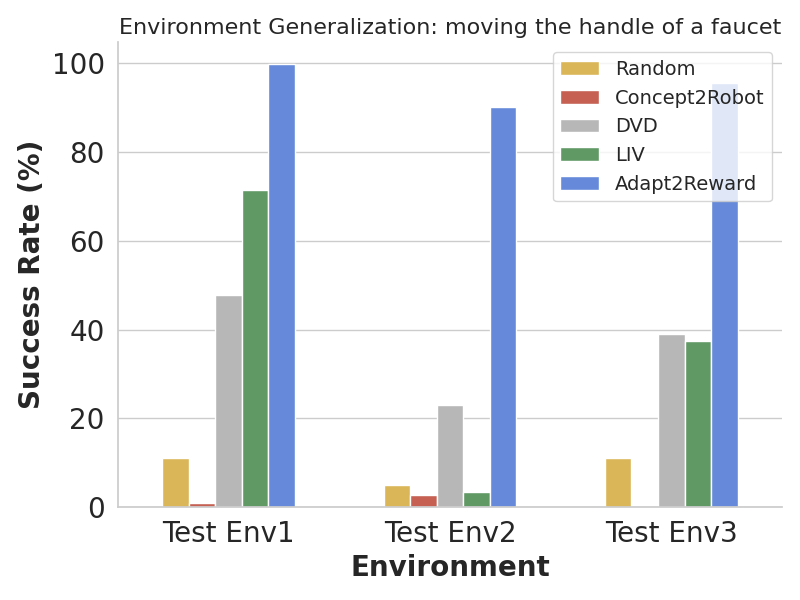}
\includegraphics[scale=0.3]{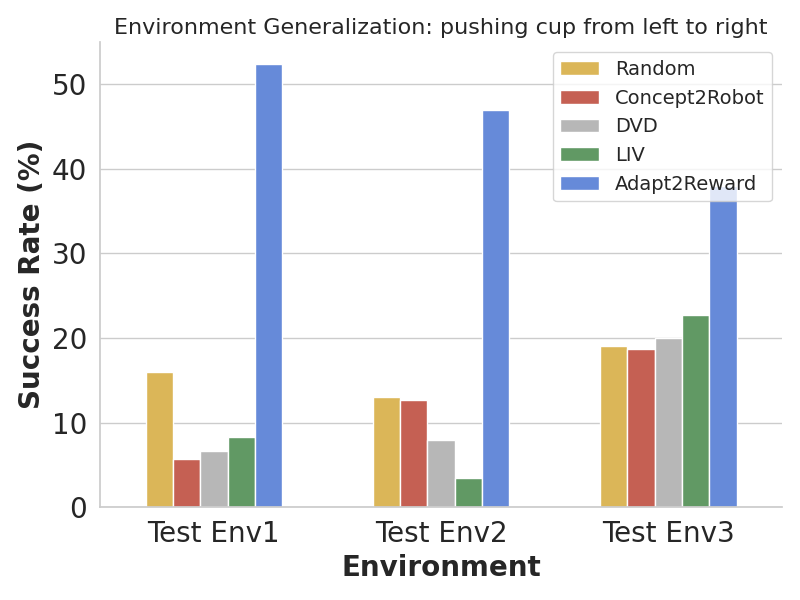}
\caption{\textbf{Environment Generalization.} We compare Adapt2Reward’s performance to DVD, Concept2Robot, and a random policy. Each bar shows the success rate for the task, computed over 3 seeds of 100 trials.
}
\label{fig_suppl_exp1}
\end{figure}
\begin{figure}[]
\centering
\includegraphics[scale=0.3]{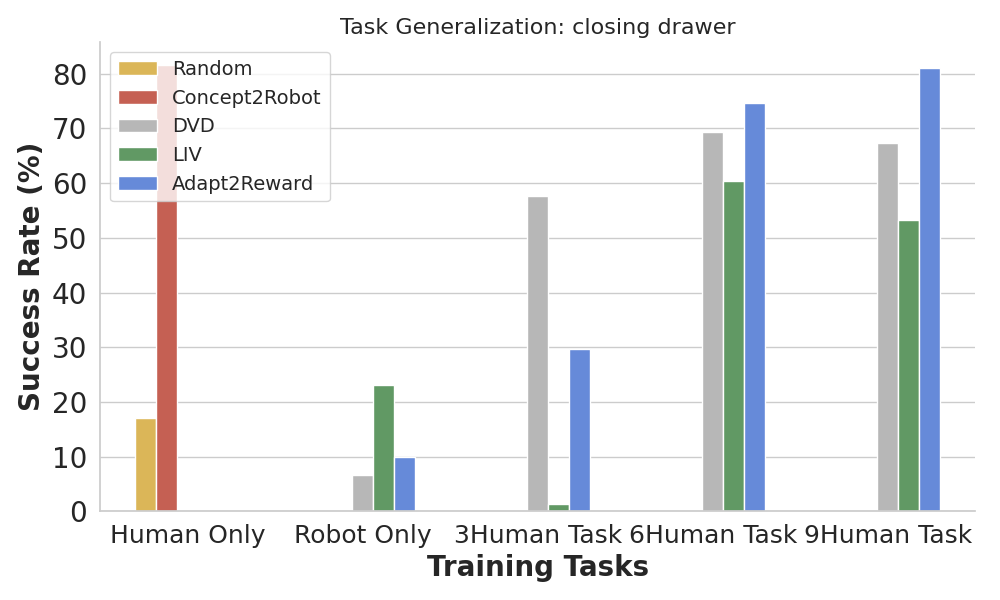}
\includegraphics[scale=0.3]{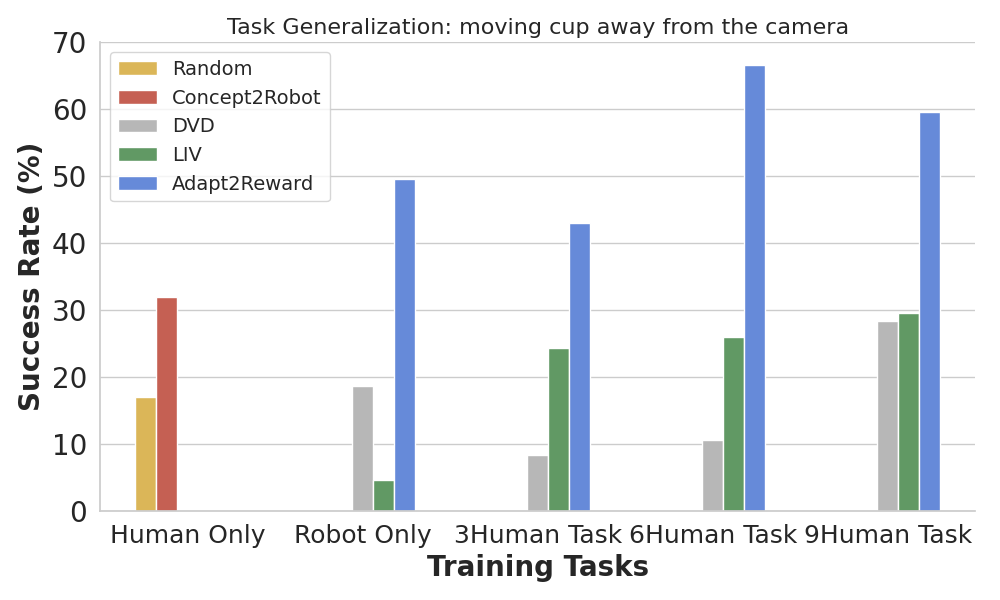}
\includegraphics[scale=0.3]{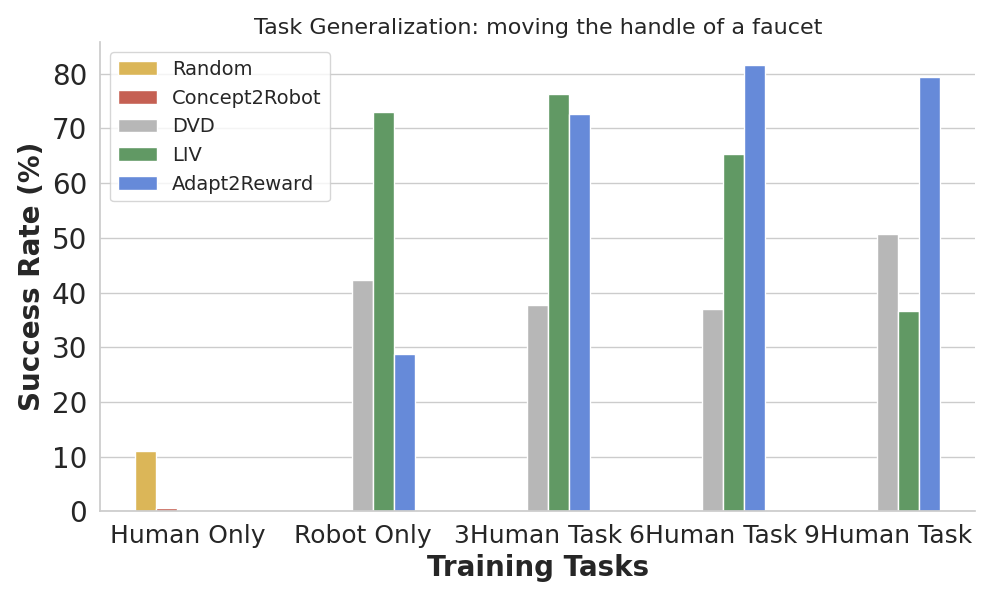}
\includegraphics[scale=0.3]{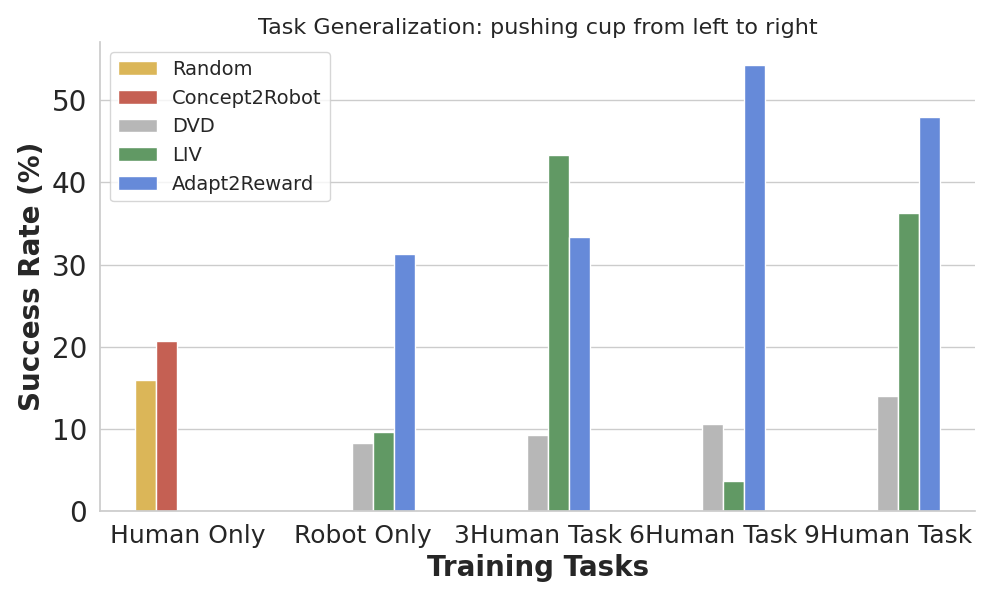}
\caption{\textbf{Task Generalization.} We compare Adapt2Reward’s performance to DVD, Concept2Robot, and a random policy. Each bar shows the success rate for the task, computed over 3 seeds of 100 trials.
}
\label{fig_suppl_exp2}
\end{figure}

\begin{figure*}[]
\centering
\includegraphics[scale=0.53]{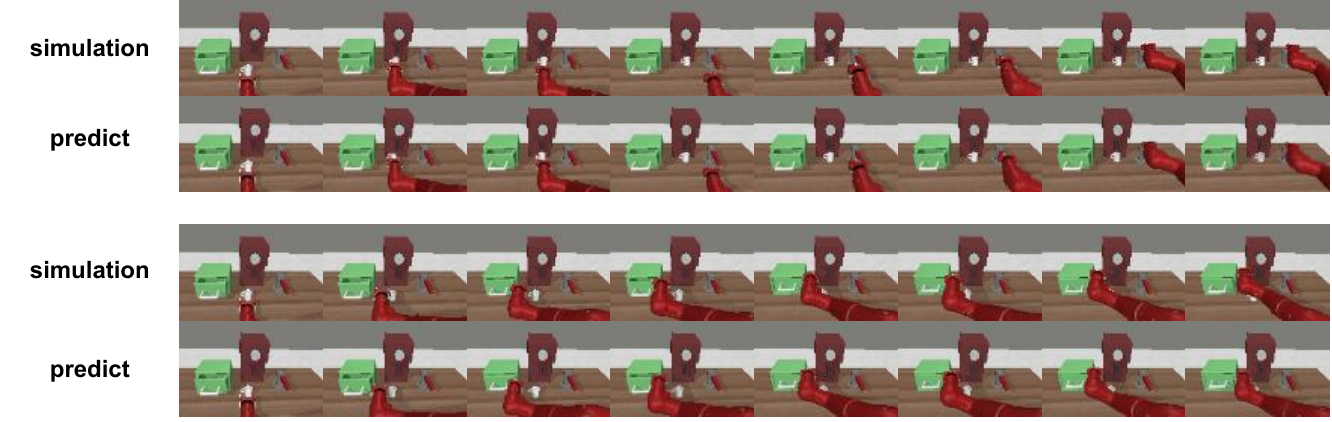}
\caption{\textbf{FitVid Performance.} We present the videos predicted by FitVid alongside actual videos from the simulation environment.  
}
\label{suppl_fivid_performance}
\end{figure*}

\begin{figure}[t]
\centering
\subfloat[Method 1]{
\includegraphics[scale=0.4]{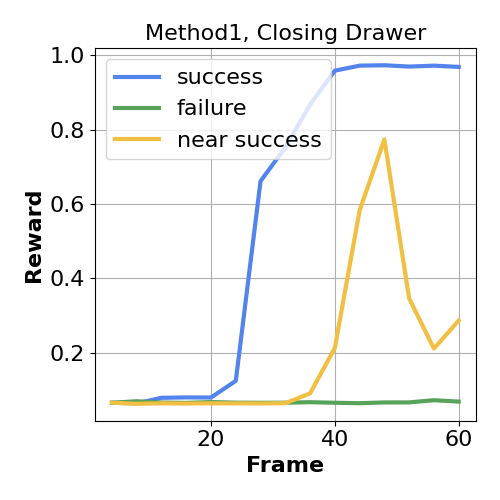}}
\subfloat[Method 2]{
\includegraphics[scale=0.4]{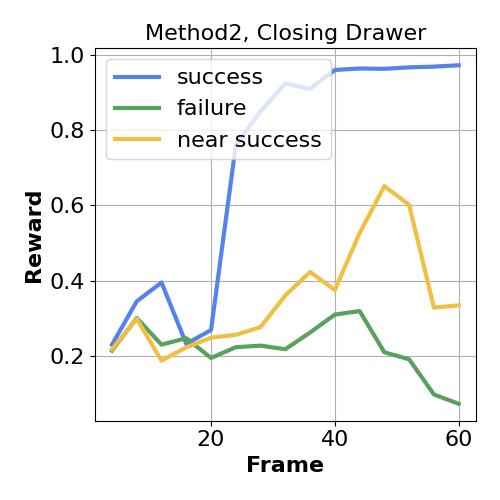}
}
\caption{\textbf{Labeling dense rewards.} We employ two distinct evaluation methods to evaluate the reliability of Adapt2Reward in assigning dense rewards and the capability of predicting future success.
}
\label{fig_suppl_dense_reward}
\end{figure}

\section{Additional Results}

\textbf{Prompt Length, Novel Objects and Vision Backbone.} In Table~\ref{ablation on failure prompt length}, our ablation study on the failure prompt length. We revealed that failure prompts of length 2 were most effective, with longer prompts potentially diverting attention from task semantics to prompt content.

In Table~\ref{change_object}, we evaluate the robot's adaptability to novel objects by introducing items it has not previously encountered, thereby challenging it to identify these objects and successfully accomplish the task. In this context, Adapt2Reward demonstrates superior performance compared to Concept2Robot, achieving a success rate that surpasses its counterpart by a significant margin of 2-52\%.

We undertook an ablation study targeting the visual encoder of the DVD model to ascertain whether the observed performance enhancement was attributable to the utilization of a distinct visual encoder in Adapt2Reward. In this process, we substituted the visual encoder in DVD with BEiT, the same encoder employed in Adapt2Reward, which had undergone pre-training on annotated labels and videos within the SSv2 dataset. The results, depicted in Figure~\ref{fig_suppl_dvd_backbone}(a), revealed a notable decline in DVD's performance following the encoder replacement. This outcome strongly suggests that the enhanced performance of Adapt2Reward cannot be attributed to the use of a different visual encoder.

\textbf{The number of robot data.} Futhermore, we explored how varying amounts of robotic data affect our model's environmental generalization ability. Figure~\ref{fig_suppl_dvd_backbone}(b) shows the success rates across environments when Adapt2Reward is trained with 100, 560, and 1000 robotic data instances. Training with 560 instances significantly enhances performance compared to just 100. However, increasing the dataset to 1000 instances from 560 yields minor improvements, hinting at a possible saturation point or limited data diversity.

\textbf{Clustering methods and t-SNE visualization.}
In addition, we explore how different clustering methods impact our experiment. Apart from K-means discussed in the main text, Figure~\ref{fig_suppl_dvd_backbone}(c) presents the results of Adapt2Reward trained using spectral clustering across four environments.
To further analyze the learned failure prompts, Figure~\ref{fig_suppl_dvd_backbone}(d) shows the t-SNE visualization of the unsuccessful videos from the "moving the handle of a faucet task" in the dataset. Different colors represent various failure prompts.

\textbf{Case Study of Failure Clusters.} In our study, we used spherical K-means clustering to iteratively update cluster centers for failure videos related to each task at the end of every training epoch. This method assigns new pseudo-labels to the failure videos, highlighting unique failure patterns for each task. As shown in Figure~\ref{suppl_task5_cluster_demos}, we set $K=3$ and display two failure videos per cluster for the ``closing drawer'' task. \textit{Cluster 1} typically includes robots moving randomly across the desktop without interacting with the drawer. \textit{Cluster 2} features robots that close the drawer and then reopen it. \textit{Cluster 3} shows robots that align with the drawer but fail to complete the closing action. These clusters shed light on the various reasons for failure in the drawer-closing task. Figure~\ref{suppl_task33_cluster_demos} shows two failure videos per cluster for the ``moving faucet handles'' task. In \textit{Cluster 1}, robots lock onto the faucet but fail to recognize the handle, leading to movements below the handle and task failure. \textit{Cluster 2} shows robots moving randomly, unable to locate the faucet. In \textit{Cluster 3}, robots correctly identify both the faucet and its handle but are unable to move the handle. Clustering failure videos reveals the different causes of failure for each task, improving the reward model's ability to distinguish between success and failure executions.

\textbf{Success Rate per Task.} In our simulation environment generalization experiments, we evaluated four tasks: 1) closing drawer, 2) moving cup away from the camera, and 3) moving the handle of a faucet, and 4) pushing cup from left to right. Figure~\ref{fig_suppl_exp1} displays the individual task success rate for Adapt2Reward, trained with seven human tasks, and compares these to previous studies.In our task generalization experiment, we explored the effect of incorporating human data on the reward function's ability to adapt to new tasks. Following the methodology in DVD~\cite{DVD}, we trained the Adapt2Reward model on robot videos from three distinct tasks within the training environment, deliberately excluding any robot data from the target tasks. The three training tasks are (1) opening drawer, (2) pushing cup from right to left, (3) poking cup so lightly that it doesn't or almost doesn't move. Figure~\ref{fig_suppl_exp2} shows the results for each of the target tasks using Adapt2Reward trained with various number of human video tasks, comparing them to previous works. The conclusion of these experiments aligns with Section 4 in the main paper.


\textbf{Performance of FitVid.} In Figure~\ref{suppl_fivid_performance}, we showcase a comparison between the videos predicted by FitVid and the corresponding actual footage from the virtual environment. Our analysis indicates that the variance between these predicted and authentic simulation videos remains within tolerable bounds. This finding effectively diminishes the concern that disparities between the predicted and real videos could substantially impact the precision of reward estimations based on these videos.

\textbf{Labeling dense rewards.}
To evaluate the reliability of Adapt2Reward in assigning dense rewards, we analyze the "closing drawer" task with successful trajectories, failed trajectories, and trajectories that were nearly successful~(closed the drawer but then opened it again). In Figure~\ref{fig_suppl_dense_reward}, we show the results of two distinct evaluation methods. The first method directly calculates the current reward using Adapt2Reward based on the frame sequence up to the current frame. The second first predicts 100 random complete trajectories from the current state using FitVid and then averages the Adapt2Reward rewards of these trajectories as the reward of the current state.

\section{Limitations and Future Works}
While our methodology has shown proficiency in generalizing to new tasks and environments through learning from a curated set of successful and failed task-specific videos, it is not without its limitations. First, it demands the preliminary collection of a subset of both successful and unsuccessful robotic videos. Second, we evaluated our method in structured simulation environments but did not test its usage in complex real-world applications where more unexpected failure scenarios occur. Third, while our approach is well-suited for straightforward, desktop-centric tasks, its efficacy is somewhat reduced for more intricate tasks that necessitate multi-stage execution processes.
In contrast, some studies leveraging Large Language Models (LLMs) like GPT-4 to craft reward functions primarily concentrate on code-based reward structures, using language as a medium for human-machine interaction. 

A promising future avenue is to exploit multimodal inputs, combining a concise set of task videos with natural language descriptions to facilitate more nuanced user interactions. This approach holds potential for more elaborate tasks, such as ``graceful walking."
Though our core strategy is straightforward and effective, there is scope for refinement. Developing methodologies to craft more sophisticated reward functions could significantly elevate success rates and bolster the capacity to tackle more intricate comparative challenges.

{
    \small
    \bibliographystyle{splncs04}
    \bibliography{main}
}